\documentclass[%
 reprint, 
 floatfix,
 amsmath,
 amssymb,
 aps, 
 physrev,
]{revtex4-2}

\usepackage{bm}
\usepackage{graphicx}
\usepackage{multirow}
\usepackage{hhline}
\usepackage{txfonts}

\begin{document}
\title{\textbf{Stage-dependent integer-binary encoding in factorization-machine black-box optimization} }%

\author{Ryo Ogawa}
\email{seonryo@keio.jp}
\affiliation{Graduate School of Science and Technology, Keio University, Kanagawa 223-8522, Japan}

\author{Mayumi Nakano}
\affiliation{Graduate School of Science and Technology, Keio University, Kanagawa 223-8522, Japan}

\author{Yuya Seki}
\affiliation{Graduate School of Science and Technology, Keio University, Kanagawa 223-8522, Japan}
\affiliation{Keio University Sustainable Quantum Artificial Intelligence Center (KSQAIC), Keio University, Tokyo 108-8345, Japan}

\author{Shu Tanaka}
\email{shu.tanaka@keio.jp}
\affiliation{Graduate School of Science and Technology, Keio University, Kanagawa 223-8522, Japan}
\affiliation{Keio University Sustainable Quantum Artificial Intelligence Center (KSQAIC), Keio University, Tokyo 108-8345, Japan}
\affiliation{Department of Applied Physics and Physico-Informatics, Keio University, Kanagawa 223-8522, Japan}
\affiliation{Human Biology-Microbiome-Quantum Research Center (WPI-Bio2Q), Keio University, Tokyo 108-8345, Japan}
\affiliation{Green Computing System Research Organization, Waseda University, Shinjuku-ku, Tokyo 162-0042, Japan}

\date{\today}

\begin{abstract}
Black-box optimization (BBO) deals with problems where objective functions lack explicit analytical forms and are expensive to evaluate.
Factorization machine with quadratic-optimization annealing (FMQA) constructs a surrogate model using a factorization machine (FM) and optimizes it with an Ising machine. 
Conventional FMQA applies a single integer-binary encoding throughout the optimization process, although the encoding best suited to surrogate learning may differ from the one best suited to Ising-machine solution search.
We propose a stage-dependent FMQA framework and derive conversion formulas between one-hot and domain-wall QUBO matrices that preserve the surrogate objective over feasible integer states up to an additive constant. 
We evaluate the OhDw variant, which employs one-hot encoding for learning and domain-wall encoding for search, on the Rastrigin function with input dimensions $N = 2$ and 5 and discretization levels $q = 61$ and 301.
Across all conditions, the dominant factor governing optimization performance is the encoding used in the learning stage, with one-hot encoding consistently yielding lower residual errors than domain-wall or binary encoding. The additional benefit of switching to domain-wall encoding for solution search is condition-dependent. 
For $N = 5$ and $q = 301$, OhDw achieves a lower residual error and solutions closer to the global optimum than one-hot-only FMQA, whereas for $N = 5$ and $q = 61$ the latter achieves a lower residual error. 
These results indicate that one-hot encoding in the learning stage is the primary performance driver and that stage-dependent encoding can provide further improvement under finer discretization.
\end{abstract}

\maketitle
\section{Introduction}
\label{sec:introduction}
A combinatorial optimization problem consists of a set of decision variables that take discrete values, an objective function that maps a combination of decision variables to a real value, and constraints that restrict the allowable values of the decision variables. 
The objective is to determine an input that minimizes the objective function while satisfying the constraints.
In combinatorial optimization problems, the size of the solution space increases exponentially with the number of variables, which makes efficient solution search difficult.
In recent years, Ising machines~\cite{mohseni2022ising,Yulianti2022QAReview,Jiang2023QABenchmark,kikuchi2025effectiveness} have attracted attention as computational devices for solving such combinatorial optimization problems. 

Ising machines are computational devices specialized for solving Ising/QUBO problems. 
They exploit various algorithms including simulated annealing (SA)~\cite{Kirkpatrick1983Optimization, Johnson1989Optimization, Johnson1991Optimization} and quantum annealing (QA)~\cite{kadowaki1998quantum,Das2008QA,Tanaka2017QSG,chakrabarti2023quantum} and can efficiently generate high-quality solutions to combinatorial optimization problems within a short computational time.
Ising machines have been reported to obtain high-quality solutions across a wide range of applications, including materials simulation
\cite{Harris2018SpinGlass,King2018Topological,Utimula2021Ionic,Sampei2023Adsorption},
transportation systems
\cite{Neukart2017Traffic,Stollenwerk2020AirTraffic,Inoue2021TrafficSignal,Mukasa2021Amusement,Marchesin2023UrbanTraffic,Kanai2024ColumnGeneration},
financial portfolio optimization
\cite{Phillipson2021Portfolio,Sakuler2025Portfolio, Takahashi2025effectiveness},
and computer-aided engineering
\cite{Endo2022PhaseField,Honda2024Truss,Xu2025Lattice}.
To solve a combinatorial optimization problem using an Ising machine, the problem is formulated as an Ising model or an equivalent quadratic unconstrained binary optimization (QUBO) model.

When the analytical form and internal structure of the objective function are unknown and only function evaluations are available, the function is referred to as a black-box (BB) function, and the corresponding optimization task is called a black-box optimization (BBO) problem.
BBO has been applied to a wide range of fields, including machine learning, physical systems, and materials design, and is an important problem in both natural and social sciences~\cite{frazier2018tutorial,lookman2019active}.
A representative approach to BBO is sequential optimization based on a surrogate model.
In this framework, a surrogate model that approximates the BB function is constructed using machine learning based on accumulated input-output data, and the next evaluation point is determined by optimizing a criterion defined based on the surrogate model.
If the predictive accuracy of the surrogate model is high, the obtained optimal solution is also expected to be a good solution for the original black-box function.
Surrogate-based optimization methods include Bayesian optimization (BO)~\cite{golovin2017google, baptista2018bayesian,oh2019combinatorial} and factorization machine with quadratic-optimization annealing (FMQA)~\cite{kitai2020designing,tamura2026black}.
FMQA is a BBO method that approximates the BB function using a factorization machine (FM)~\cite{rendle2010factorization} and performs solution search by optimizing the surrogate model with an Ising machine~\cite{kitai2020designing,tamura2026black}. 
In FMQA, by employing FM as the surrogate model, the model can be directly transformed into a QUBO form, enabling efficient optimization using an Ising machine.
FMQA has been applied to a wide range of fields, including materials design~\cite{kitai2020designing,kim2022high,kim2024wide,Lin2025ProtonIsingBBO,urushihara2023optimization,Nawa2023MTJQA}, engineering design~\cite{hida2024topology}, machine learning~\cite{tamura2024machine}, structural design~\cite{Matsumori2022QUBOStructural}, logistics~\cite{takaki2025joint}, and biomolecular systems~\cite{kikuchi2026factorization}, and has been reported to generate superior solutions compared to conventional methods such as BO, genetic algorithms, and particle swarm optimization~\cite{Matsumori2022QUBOStructural,Nawa2023MTJQA}.
In addition to these applications, methodological extensions of the FMQA framework have been explored~\cite{Minamoto2025Black, endo2025function, nakano2026swift, abe2026effectiveness, hayashi2026improving, tucs2026factorization, hama2026subsampling}.

FMQA can be applied to optimization problems involving integer and continuous variables by converting the input variables into binary variables. Integer variables can be transformed into binary variables through integer-binary encoding. Continuous variables can also be represented as binary variables by first discretizing them into integer values and then applying integer-binary encoding. Representative integer-binary encoding methods include one-hot encoding, domain-wall encoding, and binary encoding. Applications of this encoding-based framework have been reported in the optimization of photonic crystal surface-emitting laser structures, vehicle body structures, and integer-variable optimization problems~\cite{Inoue2022PhotonicQA,Kondo2025CarBodyFMQA,kikuchi2026factorization}.

Previous studies have examined how the solution search performance of Ising machines depends on the choice of integer-binary encoding.
They show that domain-wall encoding achieves lower objective function values and higher probabilities of finding optimal solutions than other encoding methods in graph coloring, flight gate assignment, and quadratic knapsack problems~\cite{kikuchi2024domainwall,chen2021performance,tamura2021performance}.
Furthermore, reference~\cite{seki2022black} demonstrated that the solution quality obtained by FMQA depends on the integer-binary encoding method.
In particular, numerical experiments reported that one-hot encoding tends to achieve lower objective function values than domain-wall encoding and binary encoding.
These results indicate that the choice of integer-binary encoding affects the optimization performance of FMQA through its influence on both the machine learning stage and the solution search stage.
The contrast between these findings, namely that domain-wall encoding is advantageous for solution search whereas one-hot encoding is advantageous for the overall performance of FMQA, suggests that the encoding suitable for the solution search stage is not necessarily identical to the encoding that maximizes the overall optimization performance of FMQA.
However, conventional FMQA uses the same encoding method in both stages.
Under such a design, the computational characteristics of each stage may not be fully exploited, and a mismatch may arise between surrogate model representation capability and solution search efficiency.
Therefore, using different encoding methods for machine learning and solution search provides a natural strategy to fully exploit the learning capability of FM and the search capability of the Ising machine.
A systematic investigation of stage-dependent encoding strategies has not been conducted.

\begin{figure*}[t]
    \centering
    \includegraphics[clip]{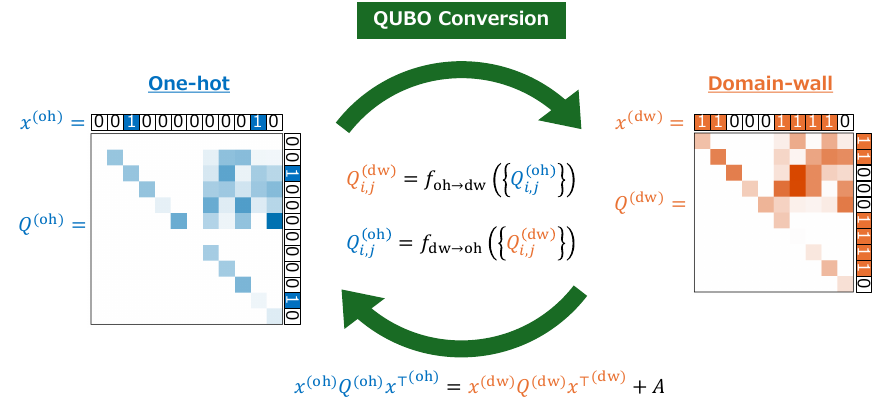}
    \caption{
    Conceptual illustration of the proposed method. The surrogate model is converted between one-hot and domain-wall encodings. By employing different integer-binary encodings in the machine learning stage and the solution search stage, the proposed method enables effective utilization of both the learning capability of the factorization machine and the search capability of the Ising machine.
    }
    \label{fig:graphical_abstract}
\end{figure*}
Based on the above considerations, this study proposes a method that employs different integer-binary encoding methods in the machine learning stage and the solution search stage when solving BBO problems with integer and continuous variables using FMQA.
The concept of the proposed method is illustrated in Fig.~\ref{fig:graphical_abstract}.
In conventional FMQA, the same integer-binary encoding is used in both the machine learning stage and the solution search stage, making it difficult to independently optimize learning performance and solution search performance.
In contrast, the proposed method first trains an FM using an encoding suitable for machine learning and then converts the resulting surrogate model into another integer-binary encoding.
This enables the use of encoding methods that are individually suited to each stage, thereby allowing both the learning capability of FM and the search capability of the Ising machine to be effectively exploited.
The objective of this study is to verify the effectiveness of the proposed method in improving the optimization performance of FMQA.

The main contributions of this study are summarized as follows.
First, we introduce an FMQA framework that employs different integer-binary encodings in the machine learning stage and the solution search stage, instead of fixing a single encoding throughout the process.
Second, we derive a mutual conversion of the QUBO matrix between one-hot and domain-wall encodings that preserves the surrogate landscape over feasible integer states, which enables the encoding used for the surrogate model to be switched between the machine learning stage and the solution search stage.
Third, we numerically decompose the effect of integer-binary encoding into its contributions to the machine learning stage and the solution search stage, and evaluate them separately.
To evaluate the proposed method, numerical experiments were conducted using the Rastrigin function~\cite{rastrigin1974systems,rudolph1990evolution}, a representative benchmark problem for black-box optimization.
The results demonstrate that the choice of integer-binary encoding in each stage significantly affects the optimization performance of FMQA, and that employing stage-dependent encodings improves performance, particularly under larger problem settings with finer discretization.

This paper is organized as follows.
Section~\ref{sec:fmqa} describes FMQA.
Section~\ref{sec:encoding} explains integer-binary encoding.
Section~\ref{sec:method} presents the proposed method.
Section~\ref{sec:setting} describes the experimental settings of the BB function, FM, and the Ising machine.
Section~\ref{sec:result} presents the optimization results of FMQA with the proposed method.
Section~\ref{sec:discussion} discusses the effects of integer-binary encoding in the learning and solution search stages.
Section~\ref{sec:conclusion} concludes this paper.

\section{FMQA}
\label{sec:fmqa}
FMQA~\cite{kitai2020designing} is a BBO method that constructs a surrogate model of a BB function using a machine learning model called an FM~\cite{rendle2010factorization} and searches for a minimum solution of the surrogate model using an Ising machine.
FMQA consists of four stages: preparation of an initial training dataset, machine learning, solution search using an Ising machine, and evaluation of the obtained solution. 
After performing the first stage once, FMQA proceeds by iteratively repeating the second to fourth stages.
The detailed procedure in each stage is described below.

In the first stage, $D_\mathrm{init}$ input vectors of $N_\mathrm{b}$-dimensional binary variables $(\boldsymbol{x}_1, \boldsymbol{x}_2, \dots, \boldsymbol{x}_{D_{\mathrm{init}}})$ are prepared, and the BB function is evaluated at each $\boldsymbol{x}_d \in \{0, 1\}^{N_\mathrm{b}} \quad (d=1, 2, \dots, D_{\mathrm{init}})$ to obtain the corresponding outputs $f_\mathrm{BB}(\boldsymbol{x}_d)$,
where $f_\mathrm{BB} : \{0,1\}^{N_\mathrm{b}} \to \mathbb{R}$ denotes the BB function.
The initial training dataset $\mathcal{D}_1$ is defined as
\begin{align}
    \label{eq:init_d}
    \mathcal{D}_1 = \{ (\boldsymbol{x}_d, y_d) \}_{d=1}^{D_\mathrm{init}},
\end{align}
where $y_d$ denotes the target value associated with input $\boldsymbol{x}_d$. In the simplest case, $y_d = f_\mathrm{BB}(\boldsymbol{x}_d)$; when output scaling is applied, $y_d$ is defined accordingly (see Sec.~\ref{sec:setting}).

In the second stage, FM is trained using the training dataset $\mathcal{D}_t$ to construct a surrogate model of the BB function, denoted by $f_\mathrm{FM}(\boldsymbol{x})$.
The index $t$ denotes the iteration number of the FMQA optimization process, and $\mathcal{D}_t$ denotes the training dataset at the $t$-th iteration.
The FM model $f_\mathrm{FM}(\boldsymbol{x})$ is expressed as
\begin{align}
    \label{eq:fm}
    f_\mathrm{FM}(\boldsymbol{x})=w_0+\sum^{N_\mathrm{b}}_{i=1}w_ix_i
    +\sum_{i=1}^{N_\mathrm{b}}\sum_{j=i+1}^{N_\mathrm{b}}\sum_{k=1}^K v_{i,k}v_{j,k}x_ix_j.
\end{align}
The integer $K \in \mathbb{N}$ denotes the rank of FM and is specified as a hyperparameter.
The parameters $w_0 \in \mathbb{R}$, $w_i \in \mathbb{R}$, and $v_{i,k} \in \mathbb{R}$ are optimized through machine learning to minimize a prescribed loss function.
In this study, the mean squared error is employed as the loss function, and the AdamW optimizer~\cite{loshchilov2017decoupled} is used for parameter updates.
The mean squared error $\mathcal{L}_{\mathrm{MSE}}$ is defined as
\begin{align}
    \label{eq:mse}
    \mathcal{L}_{\mathrm{MSE}}
    &=
    \frac{1}{|\mathcal{D}_t|}
    \sum_{(\boldsymbol{x}_d,\, y_d) \in \mathcal{D}_t}
    \left(
    f_{\mathrm{FM}}(\boldsymbol{x}_d) - y_d
    \right)^2,
\end{align}
where $|\mathcal{D}_t|$ denotes the number of data points contained in the training dataset $\mathcal{D}_t$, which is a set of distinct input-output pairs.

In the third stage, candidate solutions are generated using an Ising machine.
This is achieved by embedding the surrogate model constructed in the second stage into an Ising machine and searching for low-objective-value solutions.
To use an Ising machine, the problem is formulated as a QUBO.
The coefficients $Q_{i,j} \in \mathbb{R}$ represent the interactions between binary variables $x_i$ and $x_j$.
QUBO is expressed as follows:
\begin{align}
    \label{eq:qubo}
    \mathcal{H} = \sum_{1\leq i\leq j\leq N_\mathrm{b}} Q_{i,j} x_i x_j.
\end{align}
The QUBO can also be written in matrix form as
\begin{align}
    \label{eq:qubo_matrix}
    \mathcal{H} = \boldsymbol{x} Q \boldsymbol{x}^{\top},
\end{align}
where $\boldsymbol{x}=(x_1,\dots,x_{N_\mathrm{b}})\in\{0,1\}^{N_\mathrm{b}}$ is a row vector,
$Q\in\mathbb{R}^{{N_\mathrm{b}}\times {N_\mathrm{b}}}$ is the QUBO coefficient matrix whose
elements correspond to $Q_{i,j}$ in Eq.~\eqref{eq:qubo}, and
$\top$ denotes the transpose operation.
Since FM can be expressed in QUBO form, the model parameters $w_i$ and $v_{i,k}$ determine the QUBO coefficients $Q_{i,j}$ in Eq.~\eqref{eq:qubo}.
The constant term $w_0$ does not affect the minimization and is therefore omitted from the QUBO.
Specifically, the QUBO coefficients $Q_{i,j}$ are given by the FM parameters as follows:
\begin{equation}
    \label{eq:fmqubo}
    Q_{i,j} =
    \begin{cases}
    w_i, & i = j, \\
    \sum_{k=1}^K v_{i,k}v_{j,k}, & i < j,
    \end{cases}
\end{equation}
which clarifies the correspondence between FM and QUBO.
Here, the linear terms of the FM are absorbed into the diagonal QUBO coefficients using the relation $x_i^2 = x_i$ for $x_i \in \{0,1\}$.
Accordingly, the surrogate model $f_\mathrm{FM}(\boldsymbol{x})$ (Eq.~\eqref{eq:fm}) is embedded into the Ising machine, and a binary vector that minimizes the corresponding energy is searched.
As a result, a low-objective-value solution $\boldsymbol{x}_\mathrm{add}$ is obtained.

In the fourth stage, the obtained input $\boldsymbol{x}_\mathrm{add}$ is evaluated using the BB function to obtain the output $f_\mathrm{BB}(\boldsymbol{x}_\mathrm{add})$.
The training dataset is then updated as
\begin{align}
    \label{eq:d+1}
    \mathcal{D}_{t+1} = \mathcal{D}_t \cup \{ (\boldsymbol{x}_\mathrm{add}, y_\mathrm{add}) \}.
\end{align}
Using the updated training dataset $\mathcal{D}_{t+1}$, the machine learning procedure in the second stage is performed again.
By iteratively repeating the procedures from the second to the fourth stages, evaluated data points are accumulated sequentially.
As the number of accumulated data points increases, the surrogate model is expected to approximate the BB function more accurately, particularly in promising regions of the search space.
Consequently, candidate solutions that more accurately reflect promising regions of the search space can be generated, thereby facilitating efficient exploration toward the optimum of the BB function.
The formulation above describes FMQA for binary-variable problems. The extension to integer and continuous variables via integer-binary encoding is described in Secs.~\ref{sec:encoding} and \ref{sec:method}, where the BB function is defined on the original continuous or integer domain.

\section{Integer-binary encoding}
\label{sec:encoding}
In this section, we describe integer-binary encoding, which is used to represent integer and continuous variables in QUBO form.
Since QUBO is formulated using binary variables, integer and continuous variables must be encoded into binary variables before they can be handled in the QUBO framework.
A method for representing an integer variable using binary variables is referred to as integer-binary encoding.
We first introduce integer-binary encoding methods.
Next, we describe the discretization procedure for representing continuous variables using binary variables.
Finally, we explain the impact of encoding methods on optimization performance based on previous studies.

\subsection{Types of encoding}
\label{subsec:types}
In this subsection, we introduce representative integer-binary encoding methods, namely one-hot encoding, domain-wall encoding, and binary encoding.
Throughout this subsection, we consider an integer variable $z$ defined on $\{1,\dots,q\}$ for an arbitrary positive integer $q$, and represent it using binary variables $x$.

In one-hot encoding, $z$ is expressed using binary variables as follows:
\begin{align}
    \label{eq:one-hot}
    z=\sum_{i=1}^{q} i\,x_{i}.
\end{align}
Here, the binary vector is defined as $\boldsymbol{x} = (x_1, \dots, x_q) \in \{0, 1\}^q$.
For a given value of $z$, the binary variable with index $i=z$ is set to $1$, while all other binary variables are set to $0$.
Thus, one-hot encoding represents each feasible integer value by setting a single binary variable to $1$.
Since only one binary variable is allowed to take the value $1$, the binary variables $(x_{1}, x_{2}, \dotsc, x_{q})$ must satisfy the following encoding constraint:
\begin{align}
\label{eq:one-hot_const}
    \sum_{i=1}^{q} x_{i}=1.
\end{align}

In domain-wall encoding, $z$ is expressed as
\begin{align}
    \label{eq:domain-wall}
    z=1+\sum_{i=1}^{q-1} x_{i}.
\end{align}
Here, $\boldsymbol{x} = (x_1, \dots, x_{q-1}) \in \{0, 1\}^{q-1}$.
For a given value of $z$, the binary variables satisfying $i < z$ are set to $1$, and the remaining binary variables are set to $0$.
In domain-wall encoding, the integer value is represented by the number of consecutive binary variables taking the value $1$ from the beginning of the sequence.
Accordingly, when the binary variables are arranged in order, there exists at most one boundary at which the value changes from $1$ to $0$.
Since transitions from $0$ to $1$ must not occur in domain-wall encoding, the binary variables $(x_{1}, x_{2}, \dotsc, x_{q-1})$ must satisfy the following encoding constraint:
\begin{align}
    \label{eq:domain-wall_const}
    \sum_{i=1}^{q-2} x_{i+1}(1-x_{i})=0.
\end{align}

Using binary encoding, $z$ is expressed as
\begin{align}
    \label{eq:binary}
    z=1+\sum_{i=1}^{\lceil \log_2 q \rceil} 2^{i-1} x_{i}.
\end{align}
Here, $\boldsymbol{x} = (x_1, \dots, x_{\lceil \log_{2}{(q)} \rceil}) \in \{0, 1\}^{\lceil \log_{2}{(q)} \rceil}$, where $\lceil \log_{2}{(q)} \rceil$ denotes the smallest integer greater than or equal to $\log_2 q$.
As shown in Eq.~\eqref{eq:binary}, the integer variable is represented by expressing $z-1$ in binary form.
Since binary encoding represents integers using $\lceil \log_2(q) \rceil$ binary variables, the number of representable states can exceed the number of target integer values.
However, these assignments are decoded into integer values in the same manner by Eq.~\eqref{eq:binary}. 
Therefore, additional constraints are introduced only when the represented integer values need to be restricted to the target range.

\subsection{Discretization of continuous values}
\label{subsec:discretization}
We describe a procedure for converting a continuous variable into binary variables.
To represent a continuous variable using binary variables, its feasible range is discretized, and the discretized values are mapped to integers, after which integer-binary encoding is applied.
Specifically, let $u$ be a continuous variable whose target discretization range is $[u_{\min}, u_{\max}]$.
To map $u$ to integers in $\{1,\dots,q\}$, the interval is discretized into $q$ equally spaced points.
The discretized values are denoted by $\{u^{(z)}\}_{z=1}^{q}$.

The interval between adjacent discretized values is defined as
\begin{align}
    \Delta=\frac{u_{\max}-u_{\min}}{q-1}.
\end{align}
Using $\Delta$, the discretized values are expressed as
\begin{align}
    u^{(z)}=u_{\min}+(z-1)\Delta, \quad z=1,\dots,q.
\end{align}
Finally, by applying integer-binary encoding to the integer $z$ corresponding to each discretized value, the continuous variable is represented using binary variables.

\subsection{Effect on optimization performance}
\label{subsec:effect}
In this subsection, we briefly review the impact of integer-binary encoding methods on the optimization performance of FMQA based on previous studies.
Previous studies on combinatorial optimization problems have reported that domain-wall encoding achieves superior solution search performance in Ising machines, in terms of both objective function values and the probability of obtaining optimal solutions~\cite{kikuchi2024domainwall,chen2021performance,tamura2021performance}.
In contrast, studies on FMQA have shown that the quality of obtained solutions depends on the encoding method and that one-hot encoding tends to achieve lower objective function values than domain-wall and binary encodings for the problems considered~\cite{seki2022black}.

\section{Proposed method}
\label{sec:method}
In this section, we propose an FMQA framework that employs different integer-binary encoding methods in the machine learning and solution search stages.
As reviewed in Sec.~\ref{subsec:effect}, the encoding method affects both the estimation accuracy of the surrogate model and the solution search performance of the Ising machine, and the encoding suitable for each aspect may differ.
Based on this observation, we employ one-hot encoding in the machine learning stage and domain-wall encoding in the solution search stage, and evaluate the effectiveness of this combination.
The remainder of this section is organized as follows.
First, we introduce the QUBO matrix conversion between one-hot and domain-wall encodings.
Then, we describe the overall procedure of the proposed framework.

\subsection{Conversion of QUBO matrix}
\label{subsec:conversion}
\begin{table*}[t]
    \centering
    \caption{Parameters to be computed when converting the QUBO matrix from one-hot encoding to domain-wall encoding, the corresponding equations, the computational cost required to compute one element of the QUBO matrix, and the number of elements.}
    \begin{tabular}{cccc}
    \hline
    Parameter & Corresponding equation & Maximum computational cost per element & Number of elements\\
    \hhline{====}
    $Q_{i,j}^{l,m(\mathrm{dw})}\ (l=m\ \mathrm{and}\ i=j)$ & Eq.~\eqref{eq:Q_OhtoDw1} & $N$ & $N(q-1)$\\ 
    $Q_{i,j}^{l,m(\mathrm{dw})}\ (l < m)$ & Eq.~\eqref{eq:Q_OhtoDw2} & $1$ &$\frac{N(N-1)}{2}(q-1)^2$\\ 
    Constant term $A$  & Eq.~\eqref{eq:Q_A} & $N^2$ & $1$\\
    \hline
    \end{tabular}
    \label{table:OhNums}
\end{table*}
\begin{table*}[t]
    \centering
    \caption{Parameters to be computed when converting the QUBO matrix from domain-wall encoding to one-hot encoding, the corresponding equations, the computational cost required to compute one element of the QUBO matrix, and the number of elements.}
    \begin{tabular}{cccc}
    \hline
    Parameter & Corresponding equation & Maximum computational cost per element & Number of elements\\
    \hhline{====}
    $Q^{l,m(\mathrm{oh})'}_{i,j}\ (l=m\ \mathrm{and}\ i=j)$  & Eq.~\eqref{eq:Q_DwtoOh1} & $q$ & $Nq$\\
    $Q^{l,m(\mathrm{oh})'}_{i,j}\ (l < m)$  & Eq.~\eqref{eq:Q_DwtoOh2} & $1$ & $\frac{N(N-1)}{2}q^2$\\
    \hline
    \end{tabular}
    \label{table:DwNums}
\end{table*}
In this subsection, we describe the conversion between the QUBO matrices corresponding to one-hot encoding and domain-wall encoding.
Consider an $N$-dimensional integer input vector $\boldsymbol{z}=(z_1, z_2, \dots, z_N)$.
Each variable $z_l$ takes an integer value in the set $\{1,\dots,q\}$.
Each variable $z_l$ is converted into a binary vector by integer-binary encoding, and the resulting vectors are concatenated to form a binary vector $\boldsymbol{x}$.
The encoded vector is denoted by $\boldsymbol{x}^{(\mathrm{oh})}$ in the case of one-hot encoding and by $\boldsymbol{x}^{(\mathrm{dw})}$ in the case of domain-wall encoding.
$\boldsymbol{x}^{(\mathrm{oh})}$ is a binary vector of length $Nq$, while $\boldsymbol{x}^{(\mathrm{dw})}$ is a binary vector of length $N(q-1)$.
A QUBO is constructed for these inputs, and the corresponding outputs are denoted by
$f^{(\mathrm{oh})}(\boldsymbol{x}^{(\mathrm{oh})})$ and
$f^{(\mathrm{dw})}(\boldsymbol{x}^{(\mathrm{dw})})$.
Using the matrix representation of QUBO in Eq.~\eqref{eq:qubo_matrix}, the QUBO outputs can be expressed as
\begin{align}
    \label{eq:Hoh=}
    f^{(\mathrm{oh})}(\boldsymbol{x}^{(\mathrm{oh})})=\boldsymbol{x}^{(\mathrm{oh})}Q^\mathrm{(oh)}\boldsymbol{x}^{(\mathrm{oh})\top},
\end{align}
\begin{align}
    \label{eq:Hdw=}
    f^{(\mathrm{dw})}(\boldsymbol{x}^{(\mathrm{dw})})=\boldsymbol{x}^{(\mathrm{dw})} Q^\mathrm{(dw)}\boldsymbol{x}^{(\mathrm{dw})\top}.
\end{align}
Here, $Q^\mathrm{(oh)} \in \mathbb{R}^{Nq \times Nq}$ and $Q^\mathrm{(dw)} \in \mathbb{R}^{N(q-1) \times N(q-1)}$ denote the QUBO matrices for one-hot encoding and domain-wall encoding.
Since QUBO matrices are upper triangular, only the upper triangular elements are considered in the following derivation.
These outputs are required to agree up to an additive constant for any pair of feasible one-hot and domain-wall configurations that encode the same integer vector $\boldsymbol{z}$.
Accordingly, the following relation is required to hold:
\begin{align}
    \label{eq:Hoh=Hdw+C}
    \boldsymbol{x}^{(\mathrm{oh})}Q^\mathrm{(oh)}\boldsymbol{x}^{(\mathrm{oh})\top}=\boldsymbol{x}^{(\mathrm{dw})} Q^\mathrm{(dw)}\boldsymbol{x}^{(\mathrm{dw})\top}+A,
\end{align}
where $A \in \mathbb{R}$ is an additive constant independent of the encoded configuration.
In order to satisfy Eq.~\eqref{eq:Hoh=Hdw+C} for any feasible input configuration, the QUBO matrix under one-hot encoding, $Q^\mathrm{(oh)}$, must be appropriately transformed into the QUBO matrix under domain-wall encoding, $Q^\mathrm{(dw)}$, and vice versa.
The conversion procedure is described below.

First, we describe the conversion from $Q^{(\mathrm{oh})}$ to $Q^{(\mathrm{dw})}$. 
Let $Q_{i,j}^{l,m(\mathrm{dw})}$ denote the $(l-1)(q-1)+i$-th row and $(m-1)(q-1)+j$-th column element of $Q^{(\mathrm{dw})}$, where $l,m \in \{1,\ldots,N\}$. 
The indices $i$ and $j$ take values in $\{1,\ldots,q\}$ for one-hot encoding and in $\{1,\ldots,q-1\}$ for domain-wall encoding. 
These elements can be expressed in terms of the elements of $Q^{(\mathrm{oh})}$ as follows.
When $l=m$ and $i=j$, the relation is given by
\begin{align}
\label{eq:Q_OhtoDw1}
Q^{l,l(\mathrm{dw})}_{i,i}
&= -Q^{l,l(\mathrm{oh})}_{i,i} + Q^{l,l(\mathrm{oh})}_{i+1,i+1} \notag \\
&\quad + \sum_{a=1}^{l-1} \Big(-Q^{a,l(\mathrm{oh})}_{1,i} + Q^{a,l(\mathrm{oh})}_{1,i+1}\Big) \notag \\
&\quad + \sum_{a=l+1}^{N} \Big(-Q^{l,a(\mathrm{oh})}_{i,1} + Q^{l,a(\mathrm{oh})}_{i+1,1}\Big),
\end{align}
when $l<m$,
\begin{align}
\label{eq:Q_OhtoDw2}
Q^{l,m(\mathrm{dw})}_{i,j}
&= Q^{l,m(\mathrm{oh})}_{i,j}
  - Q^{l,m(\mathrm{oh})}_{i+1,j}
  - Q^{l,m(\mathrm{oh})}_{i,j+1}
  + Q^{l,m(\mathrm{oh})}_{i+1,j+1},
\end{align}
otherwise,
\begin{align}
\label{eq:Q_OhtoDw3}
Q^{l,m(\mathrm{dw})}_{i,j} = 0.
\end{align}
The constant term $A$ is given by
\begin{align}
\label{eq:Q_A}
A = \sum_{l=1}^{N}\sum_{m=l}^{N} Q^{l,m(\mathrm{oh})}_{1,1}.
\end{align}

Next, we describe the conversion from $Q^{(\mathrm{dw})}$ to $Q^{(\mathrm{oh})'}$.
In the following, $Q^{l,m(\mathrm{oh})'}_{i,j}$ and $Q^{l,m(\mathrm{dw})}_{i,j}$ are defined as zero when $i\leq0$ or $j\leq0$.
The $(l-1)q+i$-th row and $(m-1)q+j$-th column element of $Q^{(\mathrm{oh})'}$ is denoted by $Q_{i,j}^{l,m(\mathrm{oh})'}$.
These elements can be expressed in terms of the elements of $Q^{(\mathrm{dw})}$ as follows.
For the cases where $l=m$ and $i=j$, the relation is given as follows:
\begin{align}
\label{eq:Q_DwtoOh1}
Q^{l,l(\mathrm{oh})'}_{i,i}
&= \sum_{a=1}^{i}
\sum_{b=a}^{i}
Q^{l,l(\mathrm{dw})}_{a-1,b-1} \notag \\
&= Q^{l,l(\mathrm{oh})'}_{i-1,i-1}
+ \sum_{a=1}^{i}
Q^{l,l(\mathrm{dw})}_{a-1,i-1}.
\end{align}
For $l < m$,
\begin{align}
\label{eq:Q_DwtoOh2}
Q^{l,m(\mathrm{oh})'}_{i,j}
&= \sum_{a=1}^{i}\sum_{b=1}^{j} Q^{l,m(\mathrm{dw})}_{a-1,b-1} \notag \\
&= Q^{l,m(\mathrm{oh})'}_{i,j-1}
 + Q^{l,m(\mathrm{oh})'}_{i-1,j}
 - Q^{l,m(\mathrm{oh})'}_{i-1,j-1}
 + Q^{l,m(\mathrm{dw})}_{i-1,j-1},
\end{align}
otherwise,
\begin{align}
\label{eq:Q_DwtoOh3}
Q^{l,m(\mathrm{oh})'}_{i,j} = 0.
\end{align}
For any feasible $\boldsymbol{x}^{(\mathrm{oh})}$, the matrices $Q^\mathrm{(oh)}$ and $Q^\mathrm{(oh)'}$ satisfy
\begin{align}
\label{eq:ranQ_oh'}
\boldsymbol{x}^{(\mathrm{oh})} Q^\mathrm{(oh)}\boldsymbol{x}^{(\mathrm{oh})\top}
=
\boldsymbol{x}^{(\mathrm{oh})} Q^\mathrm{(oh)'}\boldsymbol{x}^{(\mathrm{oh})\top}
+ A .
\end{align}
That is, $Q^{(\mathrm{oh})'}$ obtained by the inverse conversion differs on feasible states only by the constant $A$, which does not affect the minimization over feasible states.

Finally, we analyze the computational complexity of these conversions.
The conversion from one-hot encoding to domain-wall encoding involves the computation of diagonal terms, off-diagonal terms, and a constant term.
The computational cost required to compute each type of element, together with the number of such elements, is summarized in Table~\ref{table:OhNums}.
Specifically, the computation in Eq.~\eqref{eq:Q_OhtoDw1} is performed $N(q-1)$ times, the computation in Eq.~\eqref{eq:Q_OhtoDw2} is performed $\frac{N(N-1)}{2}(q-1)^2$ times, and the constant term $A$ is computed once.
Accordingly, the total computational cost is given by
\begin{align}
N^2(q-1) + \frac{N(N-1)}{2}(q-1)^2 + N^2,
\end{align}
which yields a computational complexity of $\mathcal{O}(N^2 q^2)$.
Although the computational cost for the constant term $A$ is $\mathcal{O}(N^2)$, it can be ignored since it does not affect the optimization.
Similarly, the computational cost for the conversion from domain-wall encoding to one-hot encoding is summarized in Table~\ref{table:DwNums}.
Specifically, the computation of the diagonal terms in Eq.~\eqref{eq:Q_DwtoOh1} is performed $Nq^2$ times, whereas the computation of the off-diagonal terms in Eq.~\eqref{eq:Q_DwtoOh2} is performed $\frac{N(N-1)}{2}q^2$ times.
Since the computational cost per element is constant, the total computational cost is expressed as
\begin{align}
Nq^2 + \frac{N(N-1)}{2}q^2,
\end{align}
leading to the same order of complexity, $\mathcal{O}(N^2 q^2)$.
Therefore, the computational complexity required for the mutual conversion between one-hot encoding and domain-wall encoding is $\mathcal{O}(N^2 q^2)$ in both cases.

\subsection{Algorithm for the proposed method}
\label{subsec:algorithm}

\begin{figure*}[t]
    \centering
    \includegraphics[clip]{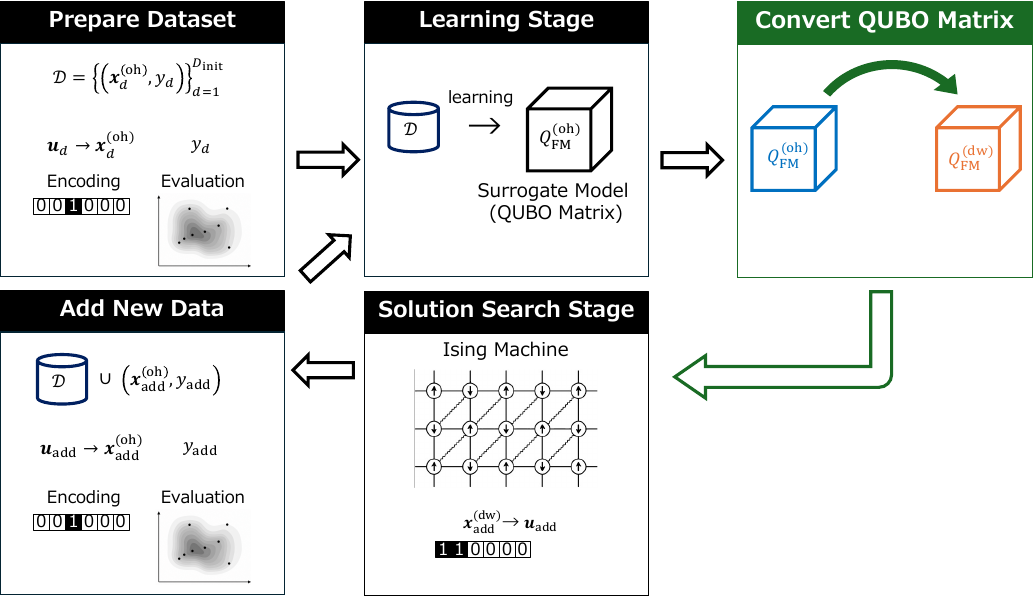}
    \caption{
    FMQA optimization procedure of the proposed method. An initial dataset is prepared and used to train an FM surrogate model. The surrogate QUBO is then converted between one-hot and domain-wall encodings according to the selected method, and optimized using an Ising machine. The obtained solution is evaluated by the BB function and added to the dataset. This process is repeated iteratively. In the OhDw method, one-hot encoding is used in the machine learning stage and domain-wall encoding is used in the solution search stage.
    }
    \label{fig:fmqa}
\end{figure*}
Based on this observation, we propose an FMQA framework that switches the integer-binary encoding between the machine learning stage and the solution search stage.
Specifically, we propose two variants: the OhDw method, which uses one-hot encoding in the machine learning stage and domain-wall encoding in the solution search stage, and the DwOh method, which adopts the opposite configuration.
As conventional baselines, we consider three single-encoding variants: the Oh method (one-hot encoding only), the Dw method (domain-wall encoding only), and the Bi method (binary encoding only).
Figure~\ref{fig:fmqa} presents the FMQA procedure in the proposed method.
In the following, we describe the procedure of the proposed method, focusing on the differences from conventional FMQA.

In the first stage, the initial training dataset is prepared in the same manner as in conventional FMQA.
Specifically, $D_\mathrm{init}$ input variables
$(\boldsymbol{u}_1, \boldsymbol{u}_2, \dots, \boldsymbol{u}_{D_{\mathrm{init}}})$
are prepared, where each input
$\boldsymbol{u}_d=(u_{d,1},u_{d,2},\dots,u_{d,N})$
consists of $N$ continuous variables taking values in the interval
$[u_\mathrm{min},u_\mathrm{max}]$.
The corresponding outputs $f_{\mathrm{BB}}(\boldsymbol{u}_d)$ are obtained by evaluating the BB function, and the initial training dataset $\mathcal{D}_1$ is constructed from these input-output pairs.

In the second stage, a surrogate model is constructed using the training dataset $\mathcal{D}_t$.
At this stage, $\boldsymbol{u}_d$ is converted into binary variables through discretization and integer-binary encoding, and machine learning is performed using the encoded variables.
The surrogate model is represented by Eq.~\eqref{eq:fm} and denoted as $f_\mathrm{FM}$.
Subsequently, the QUBO in Eq.~\eqref{eq:fmqa_hamiltonian} is constructed:
\begin{align}
    \label{eq:fmqa_hamiltonian}
    \mathcal{H} = f_\mathrm{FM}+\mu f_\mathrm{const}.
\end{align}
In practice, the QUBO coefficients derived from $f_\mathrm{FM}$ are normalized before the solution search. The normalization procedure is described in Sec.~\ref{sec:setting}.
In this expression, $f_\mathrm{const}$ represents the encoding constraint term, and $\mu \in {\mathbb{R}}$ denotes the penalty coefficient.
From Eq.~\eqref{eq:one-hot_const}, the encoding constraint term for one-hot encoding is expressed as
\begin{align}
    \label{eq:fmqa_oh_const}
f^{(\mathrm{oh})}_\mathrm{const}=\sum_{l=1}^{N}\left(\sum_{i=1}^q x^{(l)}_{i}-1\right)^2,
\end{align}
and from Eq.~\eqref{eq:domain-wall_const}, the encoding constraint term for domain-wall encoding is expressed as
\begin{align}
    \label{eq:fmqa_dw_const}
f^{\mathrm{(dw)}}_\mathrm{const}=\sum_{l=1}^{N}\left(\sum^{q-2}_{i=1}x^{(l)}_{i+1}(1-x^{(l)}_{i})\right).
\end{align}
Here, $x^{(l)}_{i}$ denotes the $i$-th binary variable used to represent the $l$-th variable.

In the third stage, the constructed QUBO is solved using an Ising machine.
In the proposed method, the QUBO matrix derived from the surrogate model $f_\mathrm{FM}$ is converted between encoding schemes before the solution search.
The encoding constraint term is not converted but is constructed directly in the target encoding.
Specifically, in the OhDw method, the surrogate QUBO matrix is converted from one-hot encoding to domain-wall encoding using Eqs.~\eqref{eq:Q_OhtoDw1}--\eqref{eq:Q_OhtoDw3}, and the domain-wall constraint term $f^{(\mathrm{dw})}_\mathrm{const}$ is added to the converted QUBO, whereas in the DwOh method, the opposite conversion is performed using Eqs.~\eqref{eq:Q_DwtoOh1}--\eqref{eq:Q_DwtoOh3}, and $f^{(\mathrm{oh})}_\mathrm{const}$ is added.
This switching of encoding between the machine learning stage and the solution search stage is the key feature of the proposed method.
The Ising machine searches for the minimum solution of $\mathcal{H}$ in Eq.~\eqref{eq:fmqa_hamiltonian}, denoted by $\boldsymbol{x}_\mathrm{add}$.
The obtained solution is represented using the integer-binary encoding employed in the solution search stage.
Solutions violating the encoding constraints are penalized by $f_\mathrm{const}$. 
When the penalty coefficient $\mu$ is sufficiently large, such solutions do not correspond to the ground state of $\mathcal{H}$.

In the fourth stage, the obtained binary solution $\boldsymbol{x}_\mathrm{add}$ is decoded into continuous-valued variables $\boldsymbol{u}_\mathrm{add}$ and evaluated using the BB function.
The resulting input-output pair is added to the training dataset, and the training dataset is updated to $\mathcal{D}_{t+1}$.
Solutions that do not satisfy the encoding constraints are discarded and replaced by a randomly generated feasible solution.
By iteratively repeating the procedures from the second stage to the fourth stage, FMQA optimization with the proposed method is performed.


\section{Settings}
\label{sec:setting}
This section describes the settings of the BB function, FM, and the Ising machine, as well as the evaluation metrics used in the simulations conducted to verify the optimization performance of FMQA with the proposed method.

\subsection{Problem}
\label{subsec:problem}
\begin{figure}[t]
    \begin{tabular}{c}
    \begin{minipage}[c]{1\hsize}
    \centering
    \includegraphics[clip]{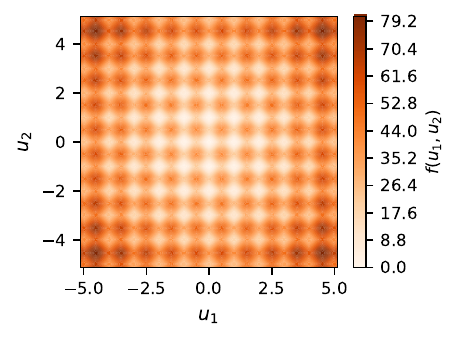}
    \end{minipage}
    \end{tabular}
    \caption{
    Rastrigin function with an input dimension of 2. 
    Colors indicate the magnitude of the output value for each input. 
    Owing to its periodic oscillatory structure, the function is multimodal and contains numerous regularly distributed local optima. 
    The global optimum is located at $\boldsymbol{u}^* = (0, 0)$, where the output value is $f_{\mathrm{BB}}(\boldsymbol{u}^*)=0$.
    }
    \label{fig:rastrigin}
\end{figure}

The Rastrigin function is a multimodal function with periodic oscillatory components, where numerous local optima are regularly distributed throughout the input space~\cite{rudolph1990evolution,rastrigin1974systems}.
For such functions, it is necessary to avoid convergence to local optima and to appropriately explore the entire search space in order to reach the optimal solution.
Therefore, the Rastrigin function has been widely used as a representative benchmark function for evaluating the global exploration capability and the ability to avoid local optima of optimization methods~\cite{demo2021supervised, Minamoto2025Black}.
The Rastrigin function is expressed as follows:
\begin{align}
\label{eq:Rastrigin}
y=f_{\mathrm{BB}}(\boldsymbol{u})
=aN+\sum_{l=1}^{N}\left(u_l^2-a\cos(2\pi u_l)\right),
\end{align}
where $\boldsymbol{u}=(u_1,\ldots,u_N)$ denotes a continuous input vector. The standard domain of the Rastrigin function is $[-5.12,5.12]^N$. 
In this study, the target discretization range is restricted to $\boldsymbol{u}\in[-3,3]^N$ so that the global optimum is contained in the representable solution set.
$N$ denotes the dimensionality of the input, and $a$ denotes a problem parameter.
In this study, $a=10$ is used.
The global minimum of the Rastrigin function is given by $\boldsymbol{u}^*=(0,0,\dots,0)$, and the corresponding function value is $f_\mathrm{BB}(\boldsymbol{u}^*)=0$.
Figure~\ref{fig:rastrigin} shows the landscape of the Rastrigin function with an input dimension of $N=2$.

In BBO problems whose outputs exhibit a wide dynamic range, constructing a highly accurate surrogate model that fits the entire input space using a limited number of input-output data points is difficult.
This is because, in learning processes aimed at minimization, data points with large output values tend to dominate the loss function, causing the model to preferentially reduce errors in high-output regions.
As a result, the accuracy of capturing the fine structure of low-objective-value regions, which are critical for identifying the optimum, is relatively degraded.
To address this issue, appropriate scaling of the output values is required.

In this study, following the prior work that applied surrogate model-based sequential BBO to the Rastrigin function~\cite{Minamoto2025Black}, the output values are scaled as follows:
\begin{align}
    \label{eq:ytoyhat}
    \bar{y} = -\exp \left(\frac{-f_\mathrm{BB}(\boldsymbol{u})}{\alpha \langle y_\mathrm{init}\rangle}\right),
\end{align}
where $\boldsymbol{u}$ denotes the input in the training dataset $\mathcal{D}$, and $\bar{y}$ denotes the scaled output value used as the training target $y_d$ in Eq.~\eqref{eq:mse}.
$\langle y_\mathrm{init}\rangle$ denotes the mean of the output values $f_\mathrm{BB}(\boldsymbol{u})$ in the initial training dataset $\mathcal{D}_1$.
$\alpha \in \mathbb{R}$ denotes a positive scaling parameter applied to the mean of the initial output values.
This scaling compresses the output values such that $\bar{y}=-1$ when $y=0$, and $\bar{y} \approx 0$ as $y \rightarrow{\infty}$.
Furthermore, the differences between scaled outputs are emphasized when $y$ is small, whereas the differences become attenuated when $y$ is large.
In other words, small values of $y$ are emphasized, while large values are compressed.
These characteristics are beneficial in the BBO framework, where the objective is to identify the region near the global minimum.
When $\langle y_\mathrm{init}\rangle$ or $\alpha$ is small, the range of emphasized output values becomes narrower, whereas larger values of $\langle y_\mathrm{init}\rangle$ or $\alpha$ broaden the emphasized range.
In this study, $\alpha=1$ is used in accordance with the prior study~\cite{Minamoto2025Black}.
This scaling requires that $y \geq 0$, which is satisfied by the BB function considered in this study.

\subsection{Parameters}
\label{subsec:parameter}
Table~\ref{table:setting} presents the settings for FMQA and the BB function.
\begin{table*}[t]
    \centering
    \caption{Settings for the BB function and FMQA}
    \begin{tabular}{c c}
    \hline
    \multicolumn{2}{c}{\textbf{BB function}} \\
    \hhline{==}
    BB function & Rastrigin function\\
    Number of input dimensions $N$ & $2,5$ \\
    Target discretization range $[u_{\min}, u_{\max}]$ & $[-3, 3]$ \\
    Number of discretization points $q$ & $61,301$ \\
    \hline
    \multicolumn{2}{c}{\rule{0pt}{2.2ex}} \\
    \hline
    \multicolumn{2}{c}{\textbf{FMQA}} \\
    \hhline{==}
    Optimizer algorithm & AdamW \\
    Learning rate & $0.5$ \\
    Learning rate schedule & Learning rate multiplied by $0.9$ every $200$ epochs \\
    Number of epochs & $2000$ \\
    Initial values of model parameters & Uniformly sampled from $[-1,1]$ \\
    Training error tolerance & $10^{-8}$ \\
    Rank of FM $K$ & $8$ \\
    Initial training dataset size  $D_\mathrm{init}$ & $64$ \\
    Number of generated solutions per search & $1$ \\
    Number of added data points & $1$ \\
    Number of iterations & $2000$ \\
    Penalty coefficient for encoding constraints $\mu$ & $1000$ \\
    Ising machine & Fixstars Amplify Annealing Engine\\
    Timeout of the Ising machine & $1000$ ms\\
    \hline
    \end{tabular}
    \label{table:setting}
\end{table*}
The number of input dimensions $N$ was set to $2$ and $5$, and the target discretization range of each continuous variable was set to $[-3,3]$.
The continuous variables were discretized into $q$ points and mapped to integers, which were subsequently transformed into binary variables using integer-binary encoding.
Increasing $q$ enables a more precise representation of the objective function.
At the same time, it increases the number of binary variables, which makes surrogate-model training and solution search using the Ising machine more challenging.
Accordingly, we considered $q=61$ and $301$ to investigate the effect of discretization resolution on optimization performance.
Under these settings, the global optimum is included in the set of representable solutions.
As described in Sec.~\ref{sec:encoding}, binary encoding theoretically allows assignments corresponding to values outside the target discretization range. 
However, no such assignments were observed in any optimization run conducted in this study. Therefore, no additional constraints were introduced to exclude them.

For FM training, the AdamW method~\cite{loshchilov2017decoupled} was employed as the optimizer. 
The default AdamW parameters proposed in Ref.~\cite{loshchilov2017decoupled} were used, namely, a weight decay of 0.01, $\beta_1=0.9$, $\beta_2=0.999$, and $\epsilon=10^{-8}$. Training was performed using full-batch optimization. At each FMQA iteration, the FM obtained in the previous iteration was used as the initial model.
The learning rate was set to $0.5$ and multiplied by a decay factor of $0.9$ every $200$ epochs. 
The maximum number of training epochs was set to $2000$. 
The hyperparameter $K$ in Eq.~\eqref{eq:fm} was set to $8$, and the initial model parameters $w_0$, $w_i$, and $v_{i,k}$ were independently sampled from a uniform distribution over $[-1,1]$. 
Training was terminated early when the training loss fell below $10^{-8}$, which was used as a convergence tolerance.

The initial training dataset size $D_{\mathrm{init}}$ was set to $64$. The initial training dataset consisted of non-duplicated feasible solutions satisfying the encoding constraints, which were uniformly sampled from the discretized search space.
In each FMQA iteration, both the number of solutions generated by the Ising machine and the number of newly added data points were set to $1$.
The total number of FMQA iterations was set to $2000$. 
If a generated solution was already contained in the training dataset, it was replaced with a randomly generated feasible solution obtained using the same procedure as that used for the initial training dataset.

The penalty coefficient for encoding constraints expressed in Eq.~\eqref{eq:fmqa_hamiltonian} was set to $\mu=1000$. Under all experimental conditions, no constraint violations were observed, indicating that this value is sufficiently large to enforce the constraints. As the Ising machine, the Fixstars Amplify Annealing Engine~\cite{FixstarsAmplify} was employed, with a timeout of $1000$ ms.

In this study, the QUBO coefficients derived from $f_{\mathrm{FM}}$ were normalized by $Q_{\mathrm{max}}$. 
Here, $Q_{\mathrm{max}}$ denotes the maximum absolute value among the elements $Q_{i,j}$ of the QUBO matrix obtained by mapping the coefficients of $f_{\mathrm{FM}}$ using Eq.~\eqref{eq:fmqubo}. 
This normalization was introduced to prevent violations of the encoding constraints from becoming energetically favorable due to excessively large model parameters during training.

\subsection{Evaluation metrics}
\label{subsec:metrix}
The optimization performance of the proposed and conventional methods is evaluated using the residual error and the norm.

Let $\hat{\boldsymbol{u}}$ denote the best solution found up to the current iteration, defined as
\begin{align}
    \hat{\boldsymbol{u}} = \operatorname*{argmin}_{\boldsymbol{u} \in \mathcal{D}_t} f_{\mathrm{BB}}(\boldsymbol{u}).
\end{align}
Both the residual error and the norm are computed with respect to this $\hat{\boldsymbol{u}}$.
The residual error represents the difference between $\hat{\boldsymbol{u}}$ obtained by FMQA and that at the global optimum $\boldsymbol{u}^*$.
The residual error $\Delta y$ is defined as follows:
\begin{align}
  \label{eq:ResidualError_hat}
  \Delta y = f_{\mathrm{BB}}(\hat{\boldsymbol{u}}) - f_{\mathrm{BB}}(\boldsymbol{u}^*).
\end{align}
Here, $f_{\mathrm{BB}}(\boldsymbol{u}^*)=0$ for the benchmark problems used in this study.
The quantity $f_{\mathrm{BB}}(\hat{\boldsymbol{u}})$ denotes the output of the BB function corresponding to the approximate solution $\hat{\boldsymbol{u}}$ obtained by FMQA.
A smaller value of $\Delta y$ indicates that the objective function value of the obtained solution is closer to that of the optimal solution $\boldsymbol{u}^*$.

The second metric is the infinity norm of the difference between the optimal solution $\boldsymbol{u}^*$ and the obtained solution $\hat{\boldsymbol{u}}$.
The infinity norm is defined as follows.
\begin{align}
  \label{eq:inf_norm_metric}
  \|\hat{\boldsymbol{u}} - \boldsymbol{u}^*\|_{\infty}
  = \max_{l \in \{1,\dots,N\}} |\hat{u}_l-u_l^*|.
\end{align}
A smaller norm indicates that the obtained solution is located closer to the optimal solution.

The Rastrigin function used in this study includes the periodic term $\cos(2\pi u_l)$, which introduces local minima at approximately unit intervals.
Accordingly, the nearest local minima to the optimal solution $\boldsymbol{u}^*=\boldsymbol{0}$ are located at approximately unit distance along each dimension.
Based on this property,
\begin{align}
\label{eq:local_zone}
  \|\boldsymbol{u}^* - \hat{\boldsymbol{u}}\|_{\infty} < 0.5,
\end{align}
indicates that the obtained solution lies in the central region around the global optimum, closer to the global optimum than to any neighboring local optimum.

\begin{figure*}[t]
    \centering
    \includegraphics[clip]{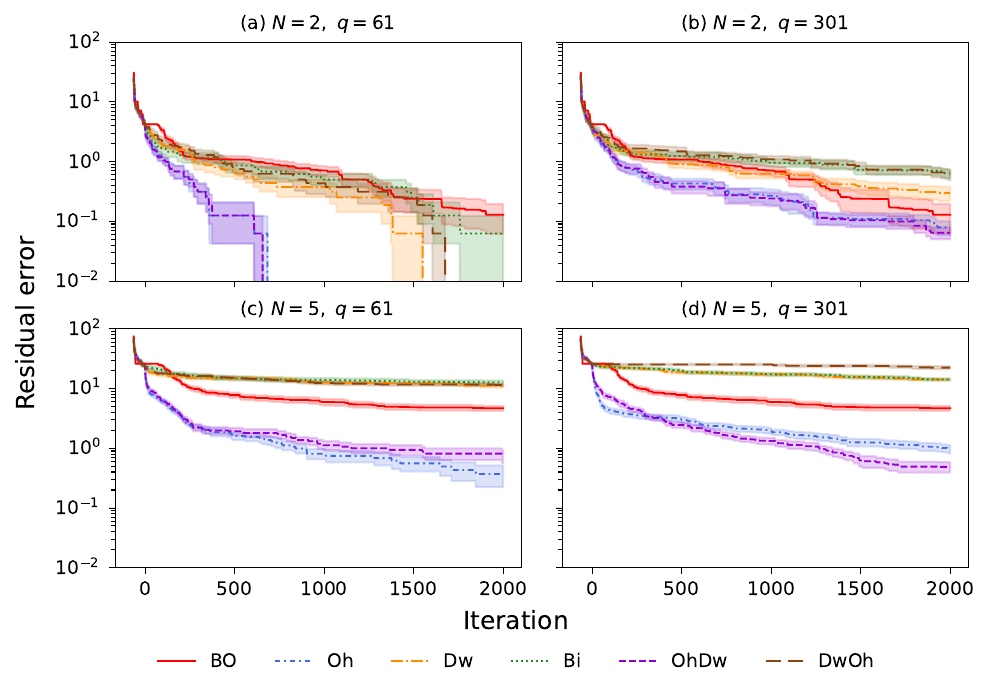}
    \caption{
    The transition of residual error values obtained during the FMQA optimization process for the Rastrigin function.
    Optimization was performed using the conventional methods represented as BO (red), Oh (blue), Dw (orange), and Bi (green), and the proposed methods represented as OhDw (purple) and DwOh (brown).
    The results are shown for (a) $N=2, q=61$, (b) $N=2, q=301$, (c) $N=5, q=61$, and (d) $N=5, q=301$.
    The lines indicate the mean values, and the shaded regions represent the standard deviations.
    }
    \label{fig:result_residualerror}
\end{figure*}

\section{Results}
\label{sec:result}
\begin{figure*}[t]
\centering
\includegraphics[clip]{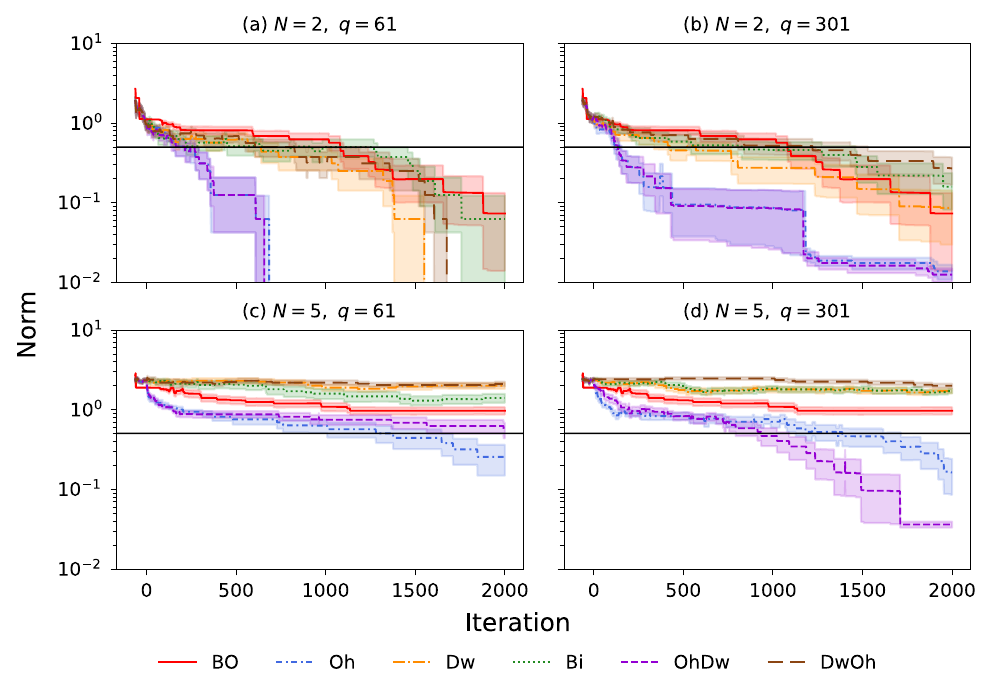}
\caption{
    The transition of norm values obtained during the FMQA optimization process for the Rastrigin function.
    Optimization was performed using the conventional methods represented as BO (red), Oh (blue), Dw (orange), and Bi (green), and the proposed methods represented as OhDw (purple) and DwOh (brown).
    The results are shown for (a) $N=2, q=61$, (b) $N=2, q=301$, (c) $N=5, q=61$, and (d) $N=5, q=301$.
    The lines indicate the mean values, and the shaded regions represent the standard deviations.
    The Rastrigin function is a multimodal function with numerous local minima. 
    In the present experimental setting, a norm value smaller than 0.5 is regarded as reaching the vicinity of the optimal solution, and the black horizontal line in the figure indicates this criterion defined by Eq.~\eqref{eq:local_zone}.
    The norm is computed for the solution $\hat{\boldsymbol{u}}$ that achieves the minimum objective function value up to each iteration, not for the solution that independently minimizes the norm.
}
\label{fig:result_norm}
\end{figure*}
The Rastrigin function is treated as the BB function, and the optimal input is searched using TPE-based sequential optimization (referred to as BO for brevity), conventional FMQA methods, namely the Oh, Dw, and Bi methods, and FMQA methods incorporating the proposed method, namely the OhDw and DwOh methods.
The input dimension of the continuous variables is set to $N=2, 5$, and the number of discretization points is set to $q=61, 301$, to evaluate the optimization performance under each condition.
In general, as $N$ and $q$ increase, the search space expands and the optimization problem becomes more challenging.

For the implementation of BO, the BBO library Optuna~\cite{akiba2019optuna} is employed.
For BO, the same 16 initial datasets were provided as starting points.
In general, BO constructs a probabilistic surrogate model of the BB function and sequentially determines the next evaluation point based on a criterion that balances exploitation and exploration.
In this study, we use the default settings of Optuna, where the Tree-structured Parzen Estimator (TPE) is adopted as the surrogate model.
The default settings were chosen to provide a standard baseline without task-specific tuning, consistent with the use of fixed hyperparameters for FMQA.
In TPE, the next evaluation point is determined by maximizing the ratio of probability densities between promising and non-promising regions.
Note that BO does not require discretization and performs optimization directly in the continuous search space.
For each setting, 16 initial datasets were generated by varying the random seed, and all methods were executed on the same set of initial datasets. This paired design ensures that the observed performance differences are not confounded with differences in the initial datasets.
This procedure reduces the dependence on the initial dataset and enables the evaluation of the average performance of each method.

Figure~\ref{fig:result_residualerror} shows the transition of the minimum residual error with respect to the number of FMQA iterations under the conditions $N=2,5$ and $q=61,301$.
The figure shows that, under all conditions, the Oh and OhDw methods, which use one-hot encoding in the machine learning stage, achieve lower residual errors than BO, Dw, Bi, and DwOh.
In particular, under the condition $N=5$, the decrease in residual error stagnates for the other methods, whereas the Oh and OhDw methods maintain low residual errors.
This result indicates that the use of one-hot encoding in the machine learning stage contributes to improving the optimization performance.

A comparison between the Oh and OhDw methods shows that, under the $N=2$ condition, no clear performance difference is observed for either $q=61$ or $q=301$.
It should be noted that, under the condition of $N=2$ and $q=61$, all FMQA methods eventually reached the global optimum, resulting in a residual error of zero. Consequently, the logarithm of the residual error diverges to $-\infty$.
In contrast, under the $N=5$ condition, a clear performance difference appears between the two methods.
In particular, for $N=5, q=61$, the Oh method achieves a slightly lower residual error than the OhDw method, whereas for $N=5, q=301$, the OhDw method achieves a lower residual error than the Oh method.
Furthermore, as $q$ increases, the search space expands and the optimization problem becomes more difficult.
The final residual error of the Oh method tends to worsen when $q$ is increased from 61 to 301.
In contrast, the OhDw method suppresses this degradation and achieves a lower final residual error at $q=301$ than at $q=61$.
Consequently, although the Oh method outperforms the OhDw method at $q=61$, the OhDw method outperforms the Oh method at $q=301$, indicating that the relative superiority of the two methods is reversed as the number of discretization points increases.
These results indicate that the OhDw method is more effective under conditions where the search space is expanded.

Figure~\ref{fig:result_norm} shows the transition of the minimum norm with respect to the number of FMQA iterations.
When the norm falls below the black line in the figure, it indicates that a solution in the vicinity of the optimal solution was obtained without falling into a local optimum.
As in the case of the residual error, the figure confirms that the Oh and OhDw methods outperform the other methods.
In particular, under the larger tested dimension $N=5$, a performance difference between the Oh and OhDw methods is observed.

Focusing on these two methods, under the condition $N=5, q=61$, the Oh method reaches the vicinity of the optimal solution, whereas the OhDw method does not fall below the black line, suggesting that solutions tend to remain near a local optimum.
However, as shown in Figure~\ref{fig:result_residualerror}(c), the residual error of the OhDw method is approximately 1.
This indicates that, although the OhDw method does not reach the optimal solution, it obtains a solution in the lowest-objective-value region among the local optima adjacent to the global optimum.
On the other hand, under the condition $N=5, q=301$, both the Oh and OhDw methods fall below the black line and obtain solutions in the vicinity of the optimal solution.
Furthermore, the OhDw method shows a smaller norm than the Oh method, confirming that it obtains a solution closer to the optimal solution.
This result indicates that, as the number of discretization points $q$ increases, the OhDw method obtains solutions in the vicinity of the global optimum, whereas at $q=61$ the obtained solutions remain near a neighboring local optimum.
This transition from a local optimum to the vicinity of the global optimum as $q$ increases is considered to be a factor contributing to the decrease in the residual error of the OhDw method.
These results suggest that, compared with the Oh method, the OhDw method may fall into a local optimum depending on the problem setting but tends to obtain solutions with lower objective function values within the same basin through solution search using the domain-wall encoding.
Therefore, although the OhDw method has a risk of stagnating at a local optimum compared with the Oh method, it can obtain solutions in deeper regions when the search proceeds effectively.

From these results, using one-hot encoding in the machine learning stage is confirmed to be important for improving the optimization performance of FMQA.
The mechanisms underlying this observation are discussed in Sec.~\ref{sec:discussion}.
Furthermore, under conditions with larger input dimensions and a larger number of discretization points, the proposed OhDw method achieves better optimization performance than the conventional Oh method.
Therefore, applying different integer-binary encoding methods to the machine learning and solution search stages in FMQA is an effective strategy for the benchmark problems considered in this study, particularly under conditions with larger input dimensions and finer discretization.

\section{Discussion}
\label{sec:discussion}
\begin{figure}[t]
    \centering
    \includegraphics[clip]{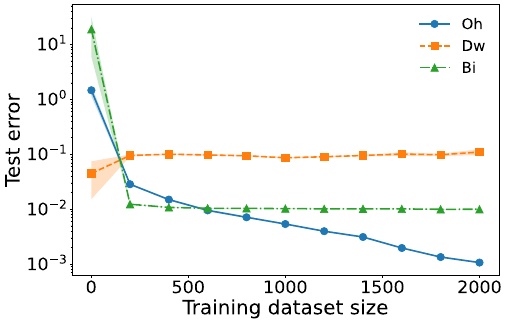}
    \caption{
    The dependence of the mean squared error (test error) on the training dataset size when learning the Rastrigin function with $N=5$ and $q=301$.
    The results obtained using one-hot encoding (blue), domain-wall encoding (orange), and binary encoding (green) are presented.
    The solid lines represent the mean values, and the shaded regions indicate the standard deviation.
    The numerical experiments were repeated 20 independent runs.
    }
    \label{fig:dis_testerror}
\end{figure}
\begin{figure*}[t]
    \centering
    \includegraphics[clip]{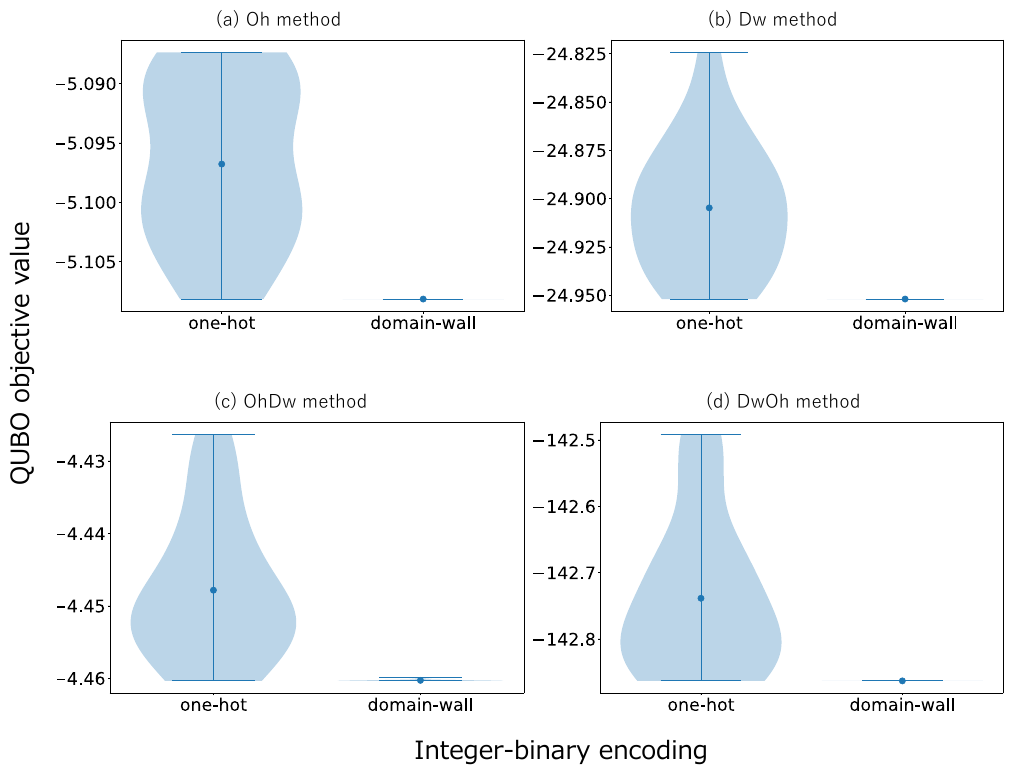}
    \caption{
    QUBO objective values obtained by Ising-machine solution search on representative surrogate models generated after FMQA optimization for the Rastrigin function with $N=5$ and $q=301$.
    One representative surrogate model was selected from each method, and both one-hot and domain-wall encodings were prepared for each model.
    (a) Oh, (b) Dw, (c) OhDw, and (d) DwOh.
    For each surrogate model, the Ising machine was executed 20 times with different random seeds.
    The plots show the distributions of the resulting QUBO objective values.
    The upper and lower bounds of the shaded regions correspond to the maximum and minimum values, respectively, and regions with a larger number of samples are shown with thicker bands.
    For each surrogate model, the constant term introduced by the QUBO conversion in Eq.~\eqref{eq:Q_A} was added to the converted QUBO so that the objective values obtained with one-hot and domain-wall encodings are directly comparable.
    }
\label{fig:dis_ising}
\end{figure*}
The proposed OhDw method was confirmed to obtain competitive or lower objective function values compared with the conventional FMQA methods, particularly under the condition with the larger tested dimension and finer discretization ($N=5$, $q=301$).
These results suggest that the encoding method optimal for the machine learning stage may differ from that suitable for the solution search stage.
In the following, we discuss the influence of integer-binary encoding methods on the estimation accuracy in the learning stage and on the search performance in the solution search stage, based on the learning mechanism of FM and the search mechanism of the Ising machine.

\subsection{Effect of encoding in the learning stage}
\label{subsec:discussion_learning}
We first investigate the influence of integer-binary encoding methods on the learning stage of FMQA, focusing on the estimation accuracy of the surrogate model.
The training dataset size was increased from 1 to 2000, corresponding to the maximum number of FMQA iterations, so that the evolution of the learning behavior can be directly associated with the optimization process.
For each training dataset size, FM was trained on input–output pairs of the Rastrigin function, where the inputs were sampled uniformly from the discretized search space.
To evaluate the generalization performance of the surrogate model near the optimal region, the training data were constructed so as not to include the global optimum or its neighboring local optima.
The model performance was evaluated using a separate test dataset consisting of 500 samples drawn uniformly from the vicinity of the global optimum defined by Eq.~\eqref{eq:local_zone}, which is particularly relevant for optimization.
The test samples were generated independently of the training dataset and did not overlap with any training samples.
The accuracy was quantified by the MSE between the FM prediction and the scaled target $\bar{y}$ defined in Eq.~\eqref{eq:ytoyhat}, evaluated on the test data.
This metric enables a systematic assessment of how the choice of encoding method affects the surrogate model accuracy, especially near the optimal solution.

Figure~\ref{fig:dis_testerror} shows the test error as a function of the training dataset size.
When machine learning was performed using one-hot encoding, the MSE decreased as the training dataset size increased, and the MSE was consistently lower than that obtained with the other encoding methods.
In contrast, for domain-wall and binary encoding, no clear decreasing trend in the MSE was observed even as the training dataset size increased.

For the condition tested ($N=5$, $q=301$), one-hot encoding yields lower test errors than domain-wall and binary encodings, indicating more effective surrogate model construction in the machine learning stage.
This characteristic is considered to be one of the factors contributing to the optimization performance achieved by the Oh and OhDw methods in FMQA.
We note that the training data in this analysis were sampled uniformly from the discretized search space, whereas in the actual FMQA optimization process the training data distribution is shaped by the solutions returned by the Ising machine and may therefore be non-uniform.

\subsection{Effect of encoding in the solution search stage}
\label{subsec:discussion_search}
We examine the influence of integer-binary encoding methods in the solution search stage.
After the completion of FMQA optimization for each method, solution search using an Ising machine was performed on the obtained surrogate models.
To isolate the effect of the encoding method in the solution search stage from that in the learning stage, each surrogate model was converted into both one-hot and domain-wall encodings, and solution search was conducted for each encoding.
The obtained solutions were evaluated using the QUBO objective value, defined as the value of the surrogate objective function $f_\mathrm{FM}(\boldsymbol{x})$ for feasible solutions satisfying the encoding constraints.
For the converted QUBO, the constant term introduced by the conversion was added to the objective value.
This enables a comparison of which encoding method can more stably generate solutions with lower QUBO objective values in the solution search stage.

Figure~\ref{fig:dis_ising} shows the outputs obtained by solution search using the Ising machine for each surrogate model.
For all methods, when solution search was performed using domain-wall encoding, the distribution of QUBO objective values was relatively narrow, and smaller QUBO objective values were generated more stably.
In contrast, when solution search was performed using one-hot encoding, cases were observed in which QUBO objective values comparable to those obtained with domain-wall encoding were achieved. 
However, the distribution was broader, and the mean value tended to be larger than that obtained with domain-wall encoding.

These results suggest that domain-wall encoding enables more stable search behavior during the solution search process of the Ising machine.
A possible explanation is that domain-wall encoding facilitates transitions between feasible states during the search process, whereas in one-hot encoding, a single-bit flip immediately produces a constraint-violating state. Consequently, transitions between feasible states under one-hot encoding require passing through constraint-violating states, which is considered to reduce search stability.
Such characteristics of domain-wall encoding in Ising-machine-based optimization have also been discussed in previous studies~\cite{kikuchi2024domainwall,chen2021performance,tamura2021performance}.

Several factors that may affect the fairness of the comparison between encoding methods in the solution search stage should be noted.
First, when the Ising machine returns a solution that is already contained in the training dataset, it is replaced by a randomly generated feasible solution. The frequency of such replacements may differ across encoding methods, which could introduce a confounding factor in the comparison.
Second, the penalty coefficient $\mu=1000$ used in this study creates a large energy barrier for constraint-violating states. Under one-hot encoding, transitions between feasible states require passing through constraint-violating states via single-bit flips, whereas domain-wall encoding allows transitions between adjacent integer values within the feasible set. The large penalty coefficient may therefore disproportionately hinder the search under one-hot encoding. A systematic investigation of the sensitivity to $\mu$ is an important direction for future work.
Third, the number of nonzero QUBO coefficients differs substantially between one-hot and domain-wall encodings due to the structure of the respective constraint terms, which may affect the solver performance independently of the encoding itself.
Fourth, under one-hot encoding, the intra-block off-diagonal products $x_i x_j$ ($i \neq j$ within the same integer-variable block) vanish on all feasible training samples. Consequently, the corresponding FM coefficients are not uniquely determined by the training data. Because the Oh and OhDw methods share the same FM model trained under one-hot encoding, this ambiguity does not affect their mutual comparison. However, when comparing surrogate models trained under different encodings (e.g., Oh versus Dw), the gauge freedom may influence the comparison through differences in the surrogate energy on infeasible states visited during the search.
Finally, the present study compared the proposed method against TPE-based optimization and single-encoding FMQA variants. Comparisons with other baseline methods, such as random search, CMA-ES, and Gaussian-process-based Bayesian optimization, remain as future work.

\section{Conclusion}
\label{sec:conclusion}
In this study, we aimed to improve the optimization performance of FMQA for BBO problems with integer and continuous variables.
We proposed a novel method for FMQA that employs different integer-binary encoding methods in the machine learning stage and in the solution search stage using an Ising machine.
To realize this framework, we derived a conversion of QUBO matrices from one-hot encoding to domain-wall encoding that preserves the surrogate landscape over feasible integer states.
In numerical experiments, we performed optimization on the Rastrigin function, a representative multimodal benchmark in BB optimization, and compared the performance of the conventional and proposed methods.
The results showed that the method employing one-hot encoding in the learning stage and domain-wall encoding in the solution search stage (OhDw method) achieved smaller residual errors and explored the vicinity of the optimum more accurately under conditions with the larger tested dimension and a large number of discretization points.
Furthermore, under the condition with the larger tested dimension and finer discretization ($N=5$, $q=301$), the proposed OhDw method, which combines one-hot encoding for learning with domain-wall encoding for solution search, achieved lower residual errors than the conventional Oh method and obtained solutions closer to the global optimum.
A key finding of this study is that integer-binary encoding methods are not merely representational choices in FMQA but design elements that serve distinct roles in the learning and solution search stages.
That is, encoding methods should not be fixed uniformly; instead, selecting them according to each stage of the optimization process can lead to improved performance.

On the other hand, several issues remain in this study.
Although similar tendencies were confirmed for the Rosenbrock function, as shown in Appendix~\ref{sec:appendix_rosenbrock}, the proposed method has been evaluated on only two benchmark functions. 
Evaluation on a wider variety of benchmark functions, including higher-dimensional problems and functions with different landscape structures, is necessary to assess the generality of the proposed method.
In addition, investigating the applicability of the proposed method to design optimization problems closer to real-world applications, such as materials design and structural optimization, is an important next step.
Furthermore, a systematic analysis of the sensitivity to the penalty coefficient $\mu$ and the effect of QUBO density differences between encoding methods would strengthen the conclusions regarding the search-stage advantage of domain-wall encoding.

\begin{acknowledgments}
This work was partially supported by the Japan Society for the Promotion of Science (JSPS) KAKENHI (Grant Numbers JP23H05447, JP25K07172), the Council for Science, Technology, and Innovation (CSTI) through the Cross-ministerial Strategic Innovation Promotion Program (SIP), ``Promoting the application of advanced quantum technology platforms to social issues'' (Funding agency: QST), Japan Science and Technology Agency (JST) (Grant Number JPMJPF2221). In addition, this paper is partially based on results obtained from a project, JPNP25014, commissioned by the New Energy and Industrial Technology Development Organization (NEDO).
S.~Tanaka wishes to express their gratitude to the World Premier International Research Center Initiative (WPI), MEXT, Japan, for their support of the Human Biology-Microbiome-Quantum Research Center (Bio2Q).
The computations in this work were partially performed using the facilities of the Supercomputer Center, the Institute for Solid State Physics, The University of Tokyo.
Also, M.~Nakano would like to express his sincere gratitude to The SATOMI Scholarship Foundation for their financial support.
\end{acknowledgments}

\appendix
\section{Results for the Rosenbrock Function}
\label{sec:appendix_rosenbrock}
\begin{figure*}[t]
    \centering
    \includegraphics[clip]{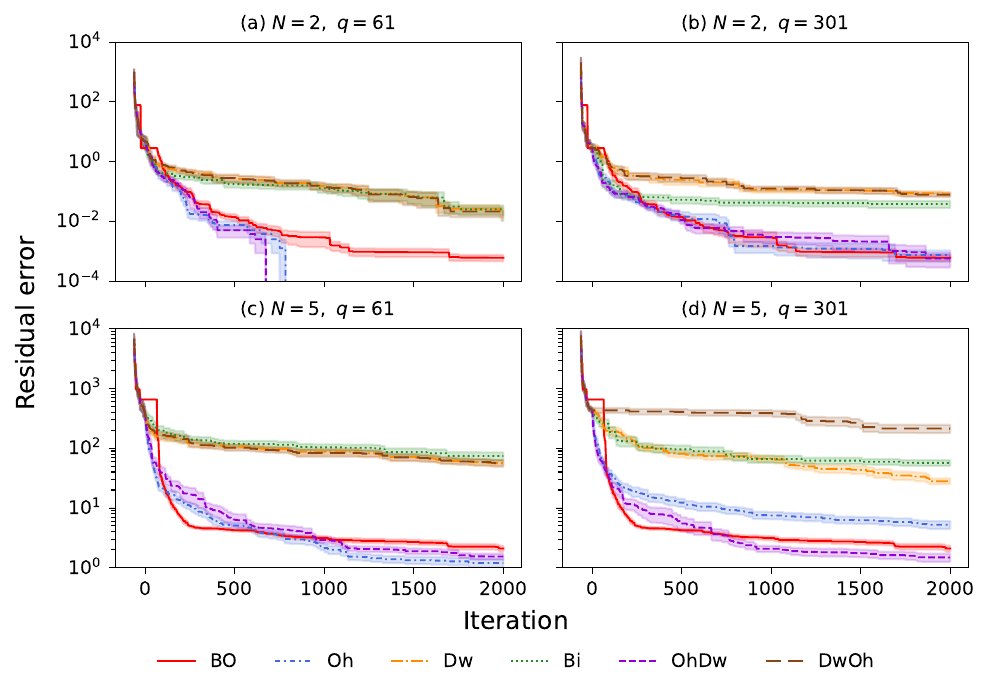}
    \caption{
    The transition of residual error values obtained during the FMQA optimization process for the Rosenbrock function.
    Optimization was performed using the conventional methods represented as BO (red), Oh (blue), Dw (orange), and Bi (green), and the proposed methods represented as OhDw (purple) and DwOh (brown).
    The results are shown for (a) $N=2, q=61$, (b) $N=2, q=301$, (c) $N=5, q=61$, and (d) $N=5, q=301$.
    The solid lines indicate the mean values, and the shaded regions represent the standard deviations.
    }
    \label{fig:Rose_result_residualerror}
\end{figure*}
\begin{figure*}[t]
\centering
\includegraphics[clip]{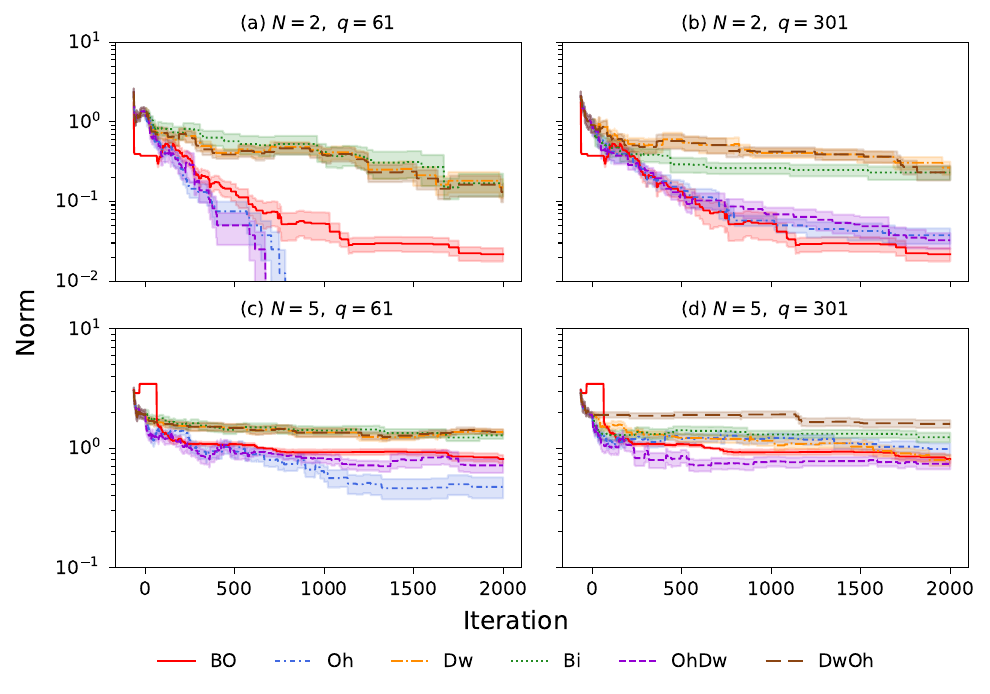}
\caption{
    The transition of norm values obtained during the FMQA optimization process for the Rosenbrock function.
    Optimization was performed using the conventional methods represented as BO (red), Oh (blue), Dw (orange), and Bi (green), and the proposed methods represented as OhDw (purple) and DwOh (brown).
    The results are shown for (a) $N=2, q=61$, (b) $N=2, q=301$, (c) $N=5, q=61$, and (d) $N=5, q=301$.
    The solid lines indicate the mean values, and the shaded regions represent the standard deviations.
    The Rosenbrock function has a unique global minimizer and a narrow curved valley.
    Although the extended function used for $N=5$ may also possess nonglobal local minima, its landscape is substantially less regularly multimodal than that of the Rastrigin function.
}
\label{fig:Rose_result_norm}
\end{figure*}
Unlike the multimodal Rastrigin function used in the main text, the Rosenbrock function, originally proposed by Rosenbrock~\cite{rosenbrock1960automatic}, has a unique global optimum and is characterized by a narrow curved valley leading to it.
Although the extended function used in this study may also possess nonglobal local minima, its landscape is substantially less regularly multimodal than that of the Rastrigin function.
Optimization remains challenging because of the flat and highly curved valley.
Therefore, the Rosenbrock function is widely used as a benchmark for evaluating optimization methods.
In this study, it is used to assess the optimization performance of each encoding method under a landscape structure different from that of the Rastrigin function.

The Rosenbrock function is expressed as follows:
\begin{align}
    \label{eq:Rosenbrock}
    f_{\mathrm{BB}}(u_1,u_2,\cdots,u_N)=\sum_{l=1}^{N-1}\left((a-u_l)^2+b(u_{l+1}-u_l^2)^2\right),
\end{align}
where $u_l$ denotes a continuous variable, $N$ denotes the input dimension, and $a$ and $b$ are problem parameters.
In this study, $a=1$ and $b=100$ are used.
The global minimum is attained at
$\boldsymbol{u}^*=(1,1,\dots,1)$,
with
$f_{\mathrm{BB}}(\boldsymbol{u}^*)=0$.

For consistency with the experiments on the Rastrigin function, the Rosenbrock function was evaluated under the same settings with $N=2,5$, $q=61,301$, and a target discretization range of $[-3,3]$.
The other experimental settings were identical to those listed in Table~\ref{table:setting}.
Figures~\ref{fig:Rose_result_residualerror} and \ref{fig:Rose_result_norm} show the optimization results for the Rosenbrock function.
Overall, the proposed OhDw method tended to achieve smaller residual errors and solution norms than conventional methods, particularly under the condition with the larger tested dimension and finer discretization ($N=5$, $q=301$).

\bibliography{reference}

\begin{thebibliography}{65}%
\makeatletter
\providecommand \@ifxundefined [1]{%
 \@ifx{#1\undefined}
}%
\providecommand \@ifnum [1]{%
 \ifnum #1\expandafter \@firstoftwo
 \else \expandafter \@secondoftwo
 \fi
}%
\providecommand \@ifx [1]{%
 \ifx #1\expandafter \@firstoftwo
 \else \expandafter \@secondoftwo
 \fi
}%
\providecommand \natexlab [1]{#1}%
\providecommand \enquote  [1]{``#1''}%
\providecommand \bibnamefont  [1]{#1}%
\providecommand \bibfnamefont [1]{#1}%
\providecommand \citenamefont [1]{#1}%
\providecommand \href@noop [0]{\@secondoftwo}%
\providecommand \href [0]{\begingroup \@sanitize@url \@href}%
\providecommand \@href[1]{\@@startlink{#1}\@@href}%
\providecommand \@@href[1]{\endgroup#1\@@endlink}%
\providecommand \@sanitize@url [0]{\catcode `\\12\catcode `\$12\catcode `\&12\catcode `\#12\catcode `\^12\catcode `\_12\catcode `\%12\relax}%
\providecommand \@@startlink[1]{}%
\providecommand \@@endlink[0]{}%
\providecommand \url  [0]{\begingroup\@sanitize@url \@url }%
\providecommand \@url [1]{\endgroup\@href {#1}{\urlprefix }}%
\providecommand \urlprefix  [0]{URL }%
\providecommand \Eprint [0]{\href }%
\providecommand \doibase [0]{https://doi.org/}%
\providecommand \selectlanguage [0]{\@gobble}%
\providecommand \bibinfo  [0]{\@secondoftwo}%
\providecommand \bibfield  [0]{\@secondoftwo}%
\providecommand \translation [1]{[#1]}%
\providecommand \BibitemOpen [0]{}%
\providecommand \bibitemStop [0]{}%
\providecommand \bibitemNoStop [0]{.\EOS\space}%
\providecommand \EOS [0]{\spacefactor3000\relax}%
\providecommand \BibitemShut  [1]{\csname bibitem#1\endcsname}%
\let\auto@bib@innerbib\@empty
\bibitem [{\citenamefont {Mohseni}\ \emph {et~al.}(2022)\citenamefont {Mohseni}, \citenamefont {McMahon},\ and\ \citenamefont {Byrnes}}]{mohseni2022ising}%
  \BibitemOpen
  \bibfield  {author} {\bibinfo {author} {\bibfnamefont {N.}~\bibnamefont {Mohseni}}, \bibinfo {author} {\bibfnamefont {P.~L.}\ \bibnamefont {McMahon}},\ and\ \bibinfo {author} {\bibfnamefont {T.}~\bibnamefont {Byrnes}},\ }\bibfield  {title} {\bibinfo {title} {{Ising} machines as hardware solvers of combinatorial optimization problems},\ }\href@noop {} {\bibfield  {journal} {\bibinfo  {journal} {Nature Reviews Physics}\ }\textbf {\bibinfo {volume} {4}},\ \bibinfo {pages} {363} (\bibinfo {year} {2022})}\BibitemShut {NoStop}%
\bibitem [{\citenamefont {Yulianti}\ and\ \citenamefont {Surendro}(2022)}]{Yulianti2022QAReview}%
  \BibitemOpen
  \bibfield  {author} {\bibinfo {author} {\bibfnamefont {L.~P.}\ \bibnamefont {Yulianti}}\ and\ \bibinfo {author} {\bibfnamefont {K.}~\bibnamefont {Surendro}},\ }\bibfield  {title} {\bibinfo {title} {Implementation of quantum annealing: A systematic review},\ }\href {https://doi.org/10.1109/ACCESS.2022.3187625} {\bibfield  {journal} {\bibinfo  {journal} {IEEE Access}\ }\textbf {\bibinfo {volume} {10}},\ \bibinfo {pages} {73156} (\bibinfo {year} {2022})}\BibitemShut {NoStop}%
\bibitem [{\citenamefont {Jiang}\ and\ \citenamefont {Chu}(2023)}]{Jiang2023QABenchmark}%
  \BibitemOpen
  \bibfield  {author} {\bibinfo {author} {\bibfnamefont {J.-R.}\ \bibnamefont {Jiang}}\ and\ \bibinfo {author} {\bibfnamefont {C.-W.}\ \bibnamefont {Chu}},\ }\bibfield  {title} {\bibinfo {title} {Classifying and benchmarking quantum annealing algorithms based on quadratic unconstrained binary optimization for solving {NP}-hard problems},\ }\href {https://doi.org/10.1109/ACCESS.2023.3315374} {\bibfield  {journal} {\bibinfo  {journal} {IEEE Access}\ }\textbf {\bibinfo {volume} {11}},\ \bibinfo {pages} {104165} (\bibinfo {year} {2023})}\BibitemShut {NoStop}%
\bibitem [{\citenamefont {Kikuchi}\ \emph {et~al.}(2025)\citenamefont {Kikuchi}, \citenamefont {Togawa},\ and\ \citenamefont {Tanaka}}]{kikuchi2025effectiveness}%
  \BibitemOpen
  \bibfield  {author} {\bibinfo {author} {\bibfnamefont {S.}~\bibnamefont {Kikuchi}}, \bibinfo {author} {\bibfnamefont {N.}~\bibnamefont {Togawa}},\ and\ \bibinfo {author} {\bibfnamefont {S.}~\bibnamefont {Tanaka}},\ }\bibfield  {title} {\bibinfo {title} {Effectiveness of hybrid optimization method for quantum annealing machines},\ }\href@noop {} {\bibfield  {journal} {\bibinfo  {journal} {arXiv preprint arXiv:2507.15544}\ } (\bibinfo {year} {2025})}\BibitemShut {NoStop}%
\bibitem [{\citenamefont {Kirkpatrick}\ \emph {et~al.}(1983)\citenamefont {Kirkpatrick}, \citenamefont {Gelatt~Jr},\ and\ \citenamefont {Vecchi}}]{Kirkpatrick1983Optimization}%
  \BibitemOpen
  \bibfield  {author} {\bibinfo {author} {\bibfnamefont {S.}~\bibnamefont {Kirkpatrick}}, \bibinfo {author} {\bibfnamefont {C.~D.}\ \bibnamefont {Gelatt~Jr}},\ and\ \bibinfo {author} {\bibfnamefont {M.~P.}\ \bibnamefont {Vecchi}},\ }\bibfield  {title} {\bibinfo {title} {Optimization by simulated annealing},\ }\href@noop {} {\bibfield  {journal} {\bibinfo  {journal} {Science}\ }\textbf {\bibinfo {volume} {220}},\ \bibinfo {pages} {671} (\bibinfo {year} {1983})}\BibitemShut {NoStop}%
\bibitem [{\citenamefont {Johnson}\ \emph {et~al.}(1989)\citenamefont {Johnson}, \citenamefont {Aragon}, \citenamefont {McGeoch},\ and\ \citenamefont {Schevon}}]{Johnson1989Optimization}%
  \BibitemOpen
  \bibfield  {author} {\bibinfo {author} {\bibfnamefont {D.~S.}\ \bibnamefont {Johnson}}, \bibinfo {author} {\bibfnamefont {C.~R.}\ \bibnamefont {Aragon}}, \bibinfo {author} {\bibfnamefont {L.~A.}\ \bibnamefont {McGeoch}},\ and\ \bibinfo {author} {\bibfnamefont {C.}~\bibnamefont {Schevon}},\ }\bibfield  {title} {\bibinfo {title} {Optimization by simulated annealing: An experimental evaluation; part {I}, graph partitioning},\ }\href@noop {} {\bibfield  {journal} {\bibinfo  {journal} {Operations Research}\ }\textbf {\bibinfo {volume} {37}},\ \bibinfo {pages} {865} (\bibinfo {year} {1989})}\BibitemShut {NoStop}%
\bibitem [{\citenamefont {Johnson}\ \emph {et~al.}(1991)\citenamefont {Johnson}, \citenamefont {Aragon}, \citenamefont {McGeoch},\ and\ \citenamefont {Schevon}}]{Johnson1991Optimization}%
  \BibitemOpen
  \bibfield  {author} {\bibinfo {author} {\bibfnamefont {D.~S.}\ \bibnamefont {Johnson}}, \bibinfo {author} {\bibfnamefont {C.~R.}\ \bibnamefont {Aragon}}, \bibinfo {author} {\bibfnamefont {L.~A.}\ \bibnamefont {McGeoch}},\ and\ \bibinfo {author} {\bibfnamefont {C.}~\bibnamefont {Schevon}},\ }\bibfield  {title} {\bibinfo {title} {Optimization by simulated annealing: an experimental evaluation; part {II}, graph coloring and number partitioning},\ }\href@noop {} {\bibfield  {journal} {\bibinfo  {journal} {Operations Research}\ }\textbf {\bibinfo {volume} {39}},\ \bibinfo {pages} {378} (\bibinfo {year} {1991})}\BibitemShut {NoStop}%
\bibitem [{\citenamefont {Kadowaki}\ and\ \citenamefont {Nishimori}(1998)}]{kadowaki1998quantum}%
  \BibitemOpen
  \bibfield  {author} {\bibinfo {author} {\bibfnamefont {T.}~\bibnamefont {Kadowaki}}\ and\ \bibinfo {author} {\bibfnamefont {H.}~\bibnamefont {Nishimori}},\ }\bibfield  {title} {\bibinfo {title} {Quantum annealing in the transverse {Ising} model},\ }\href@noop {} {\bibfield  {journal} {\bibinfo  {journal} {Physical Review E}\ }\textbf {\bibinfo {volume} {58}},\ \bibinfo {pages} {5355} (\bibinfo {year} {1998})}\BibitemShut {NoStop}%
\bibitem [{\citenamefont {Das}\ and\ \citenamefont {Chakrabarti}(2008)}]{Das2008QA}%
  \BibitemOpen
  \bibfield  {author} {\bibinfo {author} {\bibfnamefont {A.}~\bibnamefont {Das}}\ and\ \bibinfo {author} {\bibfnamefont {B.~K.}\ \bibnamefont {Chakrabarti}},\ }\bibfield  {title} {\bibinfo {title} {Colloquium: Quantum annealing and analog quantum computation},\ }\href {https://doi.org/10.1103/RevModPhys.80.1061} {\bibfield  {journal} {\bibinfo  {journal} {Reviews of Modern Physics}\ }\textbf {\bibinfo {volume} {80}},\ \bibinfo {pages} {1061} (\bibinfo {year} {2008})}\BibitemShut {NoStop}%
\bibitem [{\citenamefont {Tanaka}\ \emph {et~al.}(2017)\citenamefont {Tanaka}, \citenamefont {Tamura},\ and\ \citenamefont {Chakrabarti}}]{Tanaka2017QSG}%
  \BibitemOpen
  \bibfield  {author} {\bibinfo {author} {\bibfnamefont {S.}~\bibnamefont {Tanaka}}, \bibinfo {author} {\bibfnamefont {R.}~\bibnamefont {Tamura}},\ and\ \bibinfo {author} {\bibfnamefont {B.~K.}\ \bibnamefont {Chakrabarti}},\ }\href@noop {} {\emph {\bibinfo {title} {Quantum Spin Glasses, Annealing and Computation}}}\ (\bibinfo  {publisher} {Cambridge University Press},\ \bibinfo {address} {Cambridge, U.K.},\ \bibinfo {year} {2017})\BibitemShut {NoStop}%
\bibitem [{\citenamefont {Chakrabarti}\ \emph {et~al.}(2023)\citenamefont {Chakrabarti}, \citenamefont {Leschke}, \citenamefont {Ray}, \citenamefont {Shirai},\ and\ \citenamefont {Tanaka}}]{chakrabarti2023quantum}%
  \BibitemOpen
  \bibfield  {author} {\bibinfo {author} {\bibfnamefont {B.~K.}\ \bibnamefont {Chakrabarti}}, \bibinfo {author} {\bibfnamefont {H.}~\bibnamefont {Leschke}}, \bibinfo {author} {\bibfnamefont {P.}~\bibnamefont {Ray}}, \bibinfo {author} {\bibfnamefont {T.}~\bibnamefont {Shirai}},\ and\ \bibinfo {author} {\bibfnamefont {S.}~\bibnamefont {Tanaka}},\ }\bibfield  {title} {\bibinfo {title} {Quantum annealing and computation: challenges and perspectives},\ }\href@noop {} {\bibfield  {journal} {\bibinfo  {journal} {Philosophical Transactions of the Royal Society A}\ }\textbf {\bibinfo {volume} {381}},\ \bibinfo {pages} {20210419} (\bibinfo {year} {2023})}\BibitemShut {NoStop}%
\bibitem [{\citenamefont {Harris}\ \emph {et~al.}(2018)\citenamefont {Harris}, \citenamefont {Sato}, \citenamefont {Berkley}, \citenamefont {Reis}, \citenamefont {Altomare}, \citenamefont {Amin}, \citenamefont {Boothby}, \citenamefont {Bunyk}, \citenamefont {Deng}, \citenamefont {Enderud} \emph {et~al.}}]{Harris2018SpinGlass}%
  \BibitemOpen
  \bibfield  {author} {\bibinfo {author} {\bibfnamefont {R.}~\bibnamefont {Harris}}, \bibinfo {author} {\bibfnamefont {Y.}~\bibnamefont {Sato}}, \bibinfo {author} {\bibfnamefont {A.~J.}\ \bibnamefont {Berkley}}, \bibinfo {author} {\bibfnamefont {M.}~\bibnamefont {Reis}}, \bibinfo {author} {\bibfnamefont {F.}~\bibnamefont {Altomare}}, \bibinfo {author} {\bibfnamefont {M.~H.}\ \bibnamefont {Amin}}, \bibinfo {author} {\bibfnamefont {K.}~\bibnamefont {Boothby}}, \bibinfo {author} {\bibfnamefont {P.}~\bibnamefont {Bunyk}}, \bibinfo {author} {\bibfnamefont {C.}~\bibnamefont {Deng}}, \bibinfo {author} {\bibfnamefont {C.}~\bibnamefont {Enderud}}, \emph {et~al.},\ }\bibfield  {title} {\bibinfo {title} {Phase transitions in a programmable quantum spin glass simulator},\ }\href@noop {} {\bibfield  {journal} {\bibinfo  {journal} {Science}\ }\textbf {\bibinfo {volume} {361}},\ \bibinfo {pages} {162} (\bibinfo {year} {2018})}\BibitemShut {NoStop}%
\bibitem [{\citenamefont {King}\ \emph {et~al.}(2018)\citenamefont {King}, \citenamefont {Carrasquilla}, \citenamefont {Raymond}, \citenamefont {Ozfidan}, \citenamefont {Andriyash}, \citenamefont {Berkley}, \citenamefont {Reis}, \citenamefont {Lanting}, \citenamefont {Harris}, \citenamefont {Altomare} \emph {et~al.}}]{King2018Topological}%
  \BibitemOpen
  \bibfield  {author} {\bibinfo {author} {\bibfnamefont {A.~D.}\ \bibnamefont {King}}, \bibinfo {author} {\bibfnamefont {J.}~\bibnamefont {Carrasquilla}}, \bibinfo {author} {\bibfnamefont {J.}~\bibnamefont {Raymond}}, \bibinfo {author} {\bibfnamefont {I.}~\bibnamefont {Ozfidan}}, \bibinfo {author} {\bibfnamefont {E.}~\bibnamefont {Andriyash}}, \bibinfo {author} {\bibfnamefont {A.}~\bibnamefont {Berkley}}, \bibinfo {author} {\bibfnamefont {M.}~\bibnamefont {Reis}}, \bibinfo {author} {\bibfnamefont {T.}~\bibnamefont {Lanting}}, \bibinfo {author} {\bibfnamefont {R.}~\bibnamefont {Harris}}, \bibinfo {author} {\bibfnamefont {F.}~\bibnamefont {Altomare}}, \emph {et~al.},\ }\bibfield  {title} {\bibinfo {title} {Observation of topological phenomena in a programmable lattice of 1,800 qubits},\ }\href@noop {} {\bibfield  {journal} {\bibinfo  {journal} {Nature}\ }\textbf {\bibinfo {volume} {560}},\ \bibinfo {pages} {456} (\bibinfo {year} {2018})}\BibitemShut {NoStop}%
\bibitem [{\citenamefont {Utimula}\ \emph {et~al.}(2021)\citenamefont {Utimula}, \citenamefont {Ichibha}, \citenamefont {Prayogo}, \citenamefont {Hongo}, \citenamefont {Nakano},\ and\ \citenamefont {Maezono}}]{Utimula2021Ionic}%
  \BibitemOpen
  \bibfield  {author} {\bibinfo {author} {\bibfnamefont {K.}~\bibnamefont {Utimula}}, \bibinfo {author} {\bibfnamefont {T.}~\bibnamefont {Ichibha}}, \bibinfo {author} {\bibfnamefont {G.~I.}\ \bibnamefont {Prayogo}}, \bibinfo {author} {\bibfnamefont {K.}~\bibnamefont {Hongo}}, \bibinfo {author} {\bibfnamefont {K.}~\bibnamefont {Nakano}},\ and\ \bibinfo {author} {\bibfnamefont {R.}~\bibnamefont {Maezono}},\ }\bibfield  {title} {\bibinfo {title} {A quantum annealing approach to ionic diffusion in solids},\ }\href {https://doi.org/10.1038/s41598-021-86545-1} {\bibfield  {journal} {\bibinfo  {journal} {Scientific Reports}\ }\textbf {\bibinfo {volume} {11}},\ \bibinfo {pages} {7261} (\bibinfo {year} {2021})}\BibitemShut {NoStop}%
\bibitem [{\citenamefont {Sampei}\ \emph {et~al.}(2023)\citenamefont {Sampei}, \citenamefont {Saegusa}, \citenamefont {Chishima}, \citenamefont {Higo}, \citenamefont {Tanaka}, \citenamefont {Yayama}, \citenamefont {Nakamura}, \citenamefont {Kimura},\ and\ \citenamefont {Sekine}}]{Sampei2023Adsorption}%
  \BibitemOpen
  \bibfield  {author} {\bibinfo {author} {\bibfnamefont {H.}~\bibnamefont {Sampei}}, \bibinfo {author} {\bibfnamefont {K.}~\bibnamefont {Saegusa}}, \bibinfo {author} {\bibfnamefont {K.}~\bibnamefont {Chishima}}, \bibinfo {author} {\bibfnamefont {T.}~\bibnamefont {Higo}}, \bibinfo {author} {\bibfnamefont {S.}~\bibnamefont {Tanaka}}, \bibinfo {author} {\bibfnamefont {Y.}~\bibnamefont {Yayama}}, \bibinfo {author} {\bibfnamefont {M.}~\bibnamefont {Nakamura}}, \bibinfo {author} {\bibfnamefont {K.}~\bibnamefont {Kimura}},\ and\ \bibinfo {author} {\bibfnamefont {Y.}~\bibnamefont {Sekine}},\ }\bibfield  {title} {\bibinfo {title} {Quantum annealing boosts prediction of multimolecular adsorption on solid surfaces avoiding combinatorial explosion},\ }\href {https://doi.org/10.1021/jacsau.2c00615} {\bibfield  {journal} {\bibinfo  {journal} {JACS Au}\ }\textbf {\bibinfo {volume} {3}},\ \bibinfo {pages} {991} (\bibinfo {year} {2023})}\BibitemShut {NoStop}%
\bibitem [{\citenamefont {Neukart}\ \emph {et~al.}(2017)\citenamefont {Neukart}, \citenamefont {Compostella}, \citenamefont {Seidel}, \citenamefont {von Dollen}, \citenamefont {Yarkoni},\ and\ \citenamefont {Parney}}]{Neukart2017Traffic}%
  \BibitemOpen
  \bibfield  {author} {\bibinfo {author} {\bibfnamefont {F.}~\bibnamefont {Neukart}}, \bibinfo {author} {\bibfnamefont {G.}~\bibnamefont {Compostella}}, \bibinfo {author} {\bibfnamefont {C.}~\bibnamefont {Seidel}}, \bibinfo {author} {\bibfnamefont {D.}~\bibnamefont {von Dollen}}, \bibinfo {author} {\bibfnamefont {S.}~\bibnamefont {Yarkoni}},\ and\ \bibinfo {author} {\bibfnamefont {B.}~\bibnamefont {Parney}},\ }\bibfield  {title} {\bibinfo {title} {Traffic flow optimization using a quantum annealer},\ }\href {https://doi.org/10.3389/fict.2017.00029} {\bibfield  {journal} {\bibinfo  {journal} {Frontiers in ICT}\ }\textbf {\bibinfo {volume} {4}},\ \bibinfo {pages} {29} (\bibinfo {year} {2017})}\BibitemShut {NoStop}%
\bibitem [{\citenamefont {Stollenwerk}\ \emph {et~al.}(2020)\citenamefont {Stollenwerk}, \citenamefont {O'Gorman}, \citenamefont {Venturelli}, \citenamefont {Mandra}, \citenamefont {Rodionova}, \citenamefont {Ng}, \citenamefont {Sridhar}, \citenamefont {Rieffel},\ and\ \citenamefont {Biswas}}]{Stollenwerk2020AirTraffic}%
  \BibitemOpen
  \bibfield  {author} {\bibinfo {author} {\bibfnamefont {T.}~\bibnamefont {Stollenwerk}}, \bibinfo {author} {\bibfnamefont {B.}~\bibnamefont {O'Gorman}}, \bibinfo {author} {\bibfnamefont {D.}~\bibnamefont {Venturelli}}, \bibinfo {author} {\bibfnamefont {S.}~\bibnamefont {Mandra}}, \bibinfo {author} {\bibfnamefont {O.}~\bibnamefont {Rodionova}}, \bibinfo {author} {\bibfnamefont {H.}~\bibnamefont {Ng}}, \bibinfo {author} {\bibfnamefont {B.}~\bibnamefont {Sridhar}}, \bibinfo {author} {\bibfnamefont {E.~G.}\ \bibnamefont {Rieffel}},\ and\ \bibinfo {author} {\bibfnamefont {R.}~\bibnamefont {Biswas}},\ }\bibfield  {title} {\bibinfo {title} {Quantum annealing applied to de-conflicting optimal trajectories for air traffic management},\ }\href {https://doi.org/10.1109/TITS.2019.2891235} {\bibfield  {journal} {\bibinfo  {journal} {IEEE Transactions on Intelligent Transportation Systems}\ }\textbf {\bibinfo {volume} {21}},\ \bibinfo {pages} {285} (\bibinfo {year} {2020})}\BibitemShut {NoStop}%
\bibitem [{\citenamefont {Inoue}\ \emph {et~al.}(2021)\citenamefont {Inoue}, \citenamefont {Okada}, \citenamefont {Matsumori}, \citenamefont {Aihara},\ and\ \citenamefont {Yoshida}}]{Inoue2021TrafficSignal}%
  \BibitemOpen
  \bibfield  {author} {\bibinfo {author} {\bibfnamefont {D.}~\bibnamefont {Inoue}}, \bibinfo {author} {\bibfnamefont {A.}~\bibnamefont {Okada}}, \bibinfo {author} {\bibfnamefont {T.}~\bibnamefont {Matsumori}}, \bibinfo {author} {\bibfnamefont {K.}~\bibnamefont {Aihara}},\ and\ \bibinfo {author} {\bibfnamefont {H.}~\bibnamefont {Yoshida}},\ }\bibfield  {title} {\bibinfo {title} {Traffic signal optimization on a square lattice with quantum annealing},\ }\href {https://doi.org/10.1038/s41598-021-82771-5} {\bibfield  {journal} {\bibinfo  {journal} {Scientific Reports}\ }\textbf {\bibinfo {volume} {11}},\ \bibinfo {pages} {3303} (\bibinfo {year} {2021})}\BibitemShut {NoStop}%
\bibitem [{\citenamefont {Mukasa}\ \emph {et~al.}(2021)\citenamefont {Mukasa}, \citenamefont {Wakaizumi}, \citenamefont {Tanaka},\ and\ \citenamefont {Togawa}}]{Mukasa2021Amusement}%
  \BibitemOpen
  \bibfield  {author} {\bibinfo {author} {\bibfnamefont {Y.}~\bibnamefont {Mukasa}}, \bibinfo {author} {\bibfnamefont {K.}~\bibnamefont {Wakaizumi}}, \bibinfo {author} {\bibfnamefont {S.}~\bibnamefont {Tanaka}},\ and\ \bibinfo {author} {\bibfnamefont {N.}~\bibnamefont {Togawa}},\ }\bibfield  {title} {\bibinfo {title} {An {Ising} machine-based solver for visiting-route recommendation problems in amusement parks},\ }\href {https://doi.org/10.1587/transinf.2021EDP7064} {\bibfield  {journal} {\bibinfo  {journal} {IEICE Transactions on Information and Systems}\ }\textbf {\bibinfo {volume} {104}},\ \bibinfo {pages} {1592} (\bibinfo {year} {2021})}\BibitemShut {NoStop}%
\bibitem [{\citenamefont {Marchesin}\ \emph {et~al.}(2023)\citenamefont {Marchesin}, \citenamefont {Montrucchio}, \citenamefont {Graziano}, \citenamefont {Boella},\ and\ \citenamefont {Mondo}}]{Marchesin2023UrbanTraffic}%
  \BibitemOpen
  \bibfield  {author} {\bibinfo {author} {\bibfnamefont {A.}~\bibnamefont {Marchesin}}, \bibinfo {author} {\bibfnamefont {B.}~\bibnamefont {Montrucchio}}, \bibinfo {author} {\bibfnamefont {M.}~\bibnamefont {Graziano}}, \bibinfo {author} {\bibfnamefont {G.}~\bibnamefont {Boella}},\ and\ \bibinfo {author} {\bibfnamefont {G.}~\bibnamefont {Mondo}},\ }\bibfield  {title} {\bibinfo {title} {Improving urban traffic mobility via a versatile quantum annealing model},\ }\href {https://doi.org/10.1109/TQE.2023.3281808} {\bibfield  {journal} {\bibinfo  {journal} {IEEE Transactions on Quantum Engineering}\ }\textbf {\bibinfo {volume} {4}},\ \bibinfo {pages} {1} (\bibinfo {year} {2023})}\BibitemShut {NoStop}%
\bibitem [{\citenamefont {Kanai}\ \emph {et~al.}(2024)\citenamefont {Kanai}, \citenamefont {Yamashita}, \citenamefont {Tanahashi},\ and\ \citenamefont {Tanaka}}]{Kanai2024ColumnGeneration}%
  \BibitemOpen
  \bibfield  {author} {\bibinfo {author} {\bibfnamefont {H.}~\bibnamefont {Kanai}}, \bibinfo {author} {\bibfnamefont {M.}~\bibnamefont {Yamashita}}, \bibinfo {author} {\bibfnamefont {K.}~\bibnamefont {Tanahashi}},\ and\ \bibinfo {author} {\bibfnamefont {S.}~\bibnamefont {Tanaka}},\ }\bibfield  {title} {\bibinfo {title} {Annealing-assisted column generation for inequality-constrained combinatorial optimization problems},\ }\href {https://doi.org/10.1109/ACCESS.2024.3467821} {\bibfield  {journal} {\bibinfo  {journal} {IEEE Access}\ }\textbf {\bibinfo {volume} {12}},\ \bibinfo {pages} {157669} (\bibinfo {year} {2024})}\BibitemShut {NoStop}%
\bibitem [{\citenamefont {Phillipson}\ and\ \citenamefont {Bhatia}(2021)}]{Phillipson2021Portfolio}%
  \BibitemOpen
  \bibfield  {author} {\bibinfo {author} {\bibfnamefont {F.}~\bibnamefont {Phillipson}}\ and\ \bibinfo {author} {\bibfnamefont {H.}~\bibnamefont {Bhatia}},\ }\bibfield  {title} {\bibinfo {title} {Portfolio optimisation using the {D-Wave} quantum annealer},\ }in\ \href@noop {} {\emph {\bibinfo {booktitle} {Proceedings of the International Conference on Computational Science}}}\ (\bibinfo {address} {Krakow, Poland},\ \bibinfo {year} {2021})\ pp.\ \bibinfo {pages} {45--59}\BibitemShut {NoStop}%
\bibitem [{\citenamefont {Sakuler}\ \emph {et~al.}(2025)\citenamefont {Sakuler}, \citenamefont {Oberreuter}, \citenamefont {Aiolfi}, \citenamefont {Asproni}, \citenamefont {Roman},\ and\ \citenamefont {Schiefer}}]{Sakuler2025Portfolio}%
  \BibitemOpen
  \bibfield  {author} {\bibinfo {author} {\bibfnamefont {W.}~\bibnamefont {Sakuler}}, \bibinfo {author} {\bibfnamefont {J.~M.}\ \bibnamefont {Oberreuter}}, \bibinfo {author} {\bibfnamefont {R.}~\bibnamefont {Aiolfi}}, \bibinfo {author} {\bibfnamefont {L.}~\bibnamefont {Asproni}}, \bibinfo {author} {\bibfnamefont {B.}~\bibnamefont {Roman}},\ and\ \bibinfo {author} {\bibfnamefont {J.}~\bibnamefont {Schiefer}},\ }\bibfield  {title} {\bibinfo {title} {A real-world test of portfolio optimization with quantum annealing},\ }\href {https://doi.org/10.1007/s42484-025-00101-4} {\bibfield  {journal} {\bibinfo  {journal} {Quantum Machine Intelligence}\ }\textbf {\bibinfo {volume} {7}},\ \bibinfo {pages} {1} (\bibinfo {year} {2025})}\BibitemShut {NoStop}%
\bibitem [{\citenamefont {Takahashi}\ \emph {et~al.}(2025)\citenamefont {Takahashi}, \citenamefont {Abe}, \citenamefont {Nakamura}, \citenamefont {Hidaka}, \citenamefont {Kikuchi},\ and\ \citenamefont {Tanaka}}]{Takahashi2025effectiveness}%
  \BibitemOpen
  \bibfield  {author} {\bibinfo {author} {\bibfnamefont {K.}~\bibnamefont {Takahashi}}, \bibinfo {author} {\bibfnamefont {T.}~\bibnamefont {Abe}}, \bibinfo {author} {\bibfnamefont {Y.}~\bibnamefont {Nakamura}}, \bibinfo {author} {\bibfnamefont {R.}~\bibnamefont {Hidaka}}, \bibinfo {author} {\bibfnamefont {S.}~\bibnamefont {Kikuchi}},\ and\ \bibinfo {author} {\bibfnamefont {S.}~\bibnamefont {Tanaka}},\ }\bibfield  {title} {\bibinfo {title} {Effectiveness of cardinality-return weighted maximum independent set approach for financial portfolio optimization},\ }\href@noop {} {\bibfield  {journal} {\bibinfo  {journal} {arXiv preprint arXiv:2510.23310}\ } (\bibinfo {year} {2025})}\BibitemShut {NoStop}%
\bibitem [{\citenamefont {Endo}\ \emph {et~al.}(2022)\citenamefont {Endo}, \citenamefont {Matsuda}, \citenamefont {Tanaka},\ and\ \citenamefont {Muramatsu}}]{Endo2022PhaseField}%
  \BibitemOpen
  \bibfield  {author} {\bibinfo {author} {\bibfnamefont {K.}~\bibnamefont {Endo}}, \bibinfo {author} {\bibfnamefont {Y.}~\bibnamefont {Matsuda}}, \bibinfo {author} {\bibfnamefont {S.}~\bibnamefont {Tanaka}},\ and\ \bibinfo {author} {\bibfnamefont {M.}~\bibnamefont {Muramatsu}},\ }\bibfield  {title} {\bibinfo {title} {A phase-field model by an {Ising} machine and its application to the phase-separation structure of a diblock polymer},\ }\href {https://doi.org/10.1038/s41598-022-14871-0} {\bibfield  {journal} {\bibinfo  {journal} {Scientific Reports}\ }\textbf {\bibinfo {volume} {12}},\ \bibinfo {pages} {10794} (\bibinfo {year} {2022})}\BibitemShut {NoStop}%
\bibitem [{\citenamefont {Honda}\ \emph {et~al.}(2024)\citenamefont {Honda}, \citenamefont {Endo}, \citenamefont {Kaji}, \citenamefont {Suzuki}, \citenamefont {Matsuda}, \citenamefont {Tanaka},\ and\ \citenamefont {Muramatsu}}]{Honda2024Truss}%
  \BibitemOpen
  \bibfield  {author} {\bibinfo {author} {\bibfnamefont {R.}~\bibnamefont {Honda}}, \bibinfo {author} {\bibfnamefont {K.}~\bibnamefont {Endo}}, \bibinfo {author} {\bibfnamefont {T.}~\bibnamefont {Kaji}}, \bibinfo {author} {\bibfnamefont {Y.}~\bibnamefont {Suzuki}}, \bibinfo {author} {\bibfnamefont {Y.}~\bibnamefont {Matsuda}}, \bibinfo {author} {\bibfnamefont {S.}~\bibnamefont {Tanaka}},\ and\ \bibinfo {author} {\bibfnamefont {M.}~\bibnamefont {Muramatsu}},\ }\bibfield  {title} {\bibinfo {title} {Development of optimization method for truss structure by quantum annealing},\ }\href {https://doi.org/10.1038/s41598-024-64674-9} {\bibfield  {journal} {\bibinfo  {journal} {Scientific Reports}\ }\textbf {\bibinfo {volume} {14}},\ \bibinfo {pages} {13872} (\bibinfo {year} {2024})}\BibitemShut {NoStop}%
\bibitem [{\citenamefont {Xu}\ \emph {et~al.}(2025)\citenamefont {Xu}, \citenamefont {Shang}, \citenamefont {Kim}, \citenamefont {Lee},\ and\ \citenamefont {Luo}}]{Xu2025Lattice}%
  \BibitemOpen
  \bibfield  {author} {\bibinfo {author} {\bibfnamefont {Z.}~\bibnamefont {Xu}}, \bibinfo {author} {\bibfnamefont {W.}~\bibnamefont {Shang}}, \bibinfo {author} {\bibfnamefont {S.}~\bibnamefont {Kim}}, \bibinfo {author} {\bibfnamefont {E.}~\bibnamefont {Lee}},\ and\ \bibinfo {author} {\bibfnamefont {T.}~\bibnamefont {Luo}},\ }\bibfield  {title} {\bibinfo {title} {Quantum annealing-assisted lattice optimization},\ }\href {https://doi.org/10.1038/s41524-024-01365-2} {\bibfield  {journal} {\bibinfo  {journal} {npj Computational Materials}\ }\textbf {\bibinfo {volume} {11}},\ \bibinfo {pages} {4} (\bibinfo {year} {2025})}\BibitemShut {NoStop}%
\bibitem [{\citenamefont {Frazier}(2018)}]{frazier2018tutorial}%
  \BibitemOpen
  \bibfield  {author} {\bibinfo {author} {\bibfnamefont {P.~I.}\ \bibnamefont {Frazier}},\ }\bibfield  {title} {\bibinfo {title} {A tutorial on {Bayesian} optimization},\ }\href@noop {} {\bibfield  {journal} {\bibinfo  {journal} {arXiv preprint arXiv:1807.02811}\ } (\bibinfo {year} {2018})}\BibitemShut {NoStop}%
\bibitem [{\citenamefont {Lookman}\ \emph {et~al.}(2019)\citenamefont {Lookman}, \citenamefont {Balachandran}, \citenamefont {Xue},\ and\ \citenamefont {Yuan}}]{lookman2019active}%
  \BibitemOpen
  \bibfield  {author} {\bibinfo {author} {\bibfnamefont {T.}~\bibnamefont {Lookman}}, \bibinfo {author} {\bibfnamefont {P.~V.}\ \bibnamefont {Balachandran}}, \bibinfo {author} {\bibfnamefont {D.}~\bibnamefont {Xue}},\ and\ \bibinfo {author} {\bibfnamefont {R.}~\bibnamefont {Yuan}},\ }\bibfield  {title} {\bibinfo {title} {Active learning in materials science with emphasis on adaptive sampling using uncertainties for targeted design},\ }\href@noop {} {\bibfield  {journal} {\bibinfo  {journal} {npj Computational Materials}\ }\textbf {\bibinfo {volume} {5}},\ \bibinfo {pages} {21} (\bibinfo {year} {2019})}\BibitemShut {NoStop}%
\bibitem [{\citenamefont {Golovin}\ \emph {et~al.}(2017)\citenamefont {Golovin}, \citenamefont {Solnik}, \citenamefont {Moitra}, \citenamefont {Kochanski}, \citenamefont {Karro},\ and\ \citenamefont {Sculley}}]{golovin2017google}%
  \BibitemOpen
  \bibfield  {author} {\bibinfo {author} {\bibfnamefont {D.}~\bibnamefont {Golovin}}, \bibinfo {author} {\bibfnamefont {B.}~\bibnamefont {Solnik}}, \bibinfo {author} {\bibfnamefont {S.}~\bibnamefont {Moitra}}, \bibinfo {author} {\bibfnamefont {G.}~\bibnamefont {Kochanski}}, \bibinfo {author} {\bibfnamefont {J.}~\bibnamefont {Karro}},\ and\ \bibinfo {author} {\bibfnamefont {D.}~\bibnamefont {Sculley}},\ }\bibfield  {title} {\bibinfo {title} {Google {Vizier}: A service for black-box optimization},\ }in\ \href@noop {} {\emph {\bibinfo {booktitle} {Proceedings of the 23rd ACM SIGKDD international conference on knowledge discovery and data mining}}}\ (\bibinfo {year} {2017})\ pp.\ \bibinfo {pages} {1487--1495}\BibitemShut {NoStop}%
\bibitem [{\citenamefont {Baptista}\ and\ \citenamefont {Poloczek}(2018)}]{baptista2018bayesian}%
  \BibitemOpen
  \bibfield  {author} {\bibinfo {author} {\bibfnamefont {R.}~\bibnamefont {Baptista}}\ and\ \bibinfo {author} {\bibfnamefont {M.}~\bibnamefont {Poloczek}},\ }\bibfield  {title} {\bibinfo {title} {{Bayesian} optimization of combinatorial structures},\ }in\ \href@noop {} {\emph {\bibinfo {booktitle} {International conference on machine learning}}}\ (\bibinfo {organization} {PMLR},\ \bibinfo {year} {2018})\ pp.\ \bibinfo {pages} {462--471}\BibitemShut {NoStop}%
\bibitem [{\citenamefont {Oh}\ \emph {et~al.}(2019)\citenamefont {Oh}, \citenamefont {Tomczak}, \citenamefont {Gavves},\ and\ \citenamefont {Welling}}]{oh2019combinatorial}%
  \BibitemOpen
  \bibfield  {author} {\bibinfo {author} {\bibfnamefont {C.}~\bibnamefont {Oh}}, \bibinfo {author} {\bibfnamefont {J.}~\bibnamefont {Tomczak}}, \bibinfo {author} {\bibfnamefont {E.}~\bibnamefont {Gavves}},\ and\ \bibinfo {author} {\bibfnamefont {M.}~\bibnamefont {Welling}},\ }\bibfield  {title} {\bibinfo {title} {Combinatorial {Bayesian} optimization using the graph {Cartesian} product},\ }\href@noop {} {\bibfield  {journal} {\bibinfo  {journal} {Advances in Neural Information Processing Systems}\ }\textbf {\bibinfo {volume} {32}} (\bibinfo {year} {2019})}\BibitemShut {NoStop}%
\bibitem [{\citenamefont {Kitai}\ \emph {et~al.}(2020)\citenamefont {Kitai}, \citenamefont {Guo}, \citenamefont {Ju}, \citenamefont {Tanaka}, \citenamefont {Tsuda}, \citenamefont {Shiomi},\ and\ \citenamefont {Tamura}}]{kitai2020designing}%
  \BibitemOpen
  \bibfield  {author} {\bibinfo {author} {\bibfnamefont {K.}~\bibnamefont {Kitai}}, \bibinfo {author} {\bibfnamefont {J.}~\bibnamefont {Guo}}, \bibinfo {author} {\bibfnamefont {S.}~\bibnamefont {Ju}}, \bibinfo {author} {\bibfnamefont {S.}~\bibnamefont {Tanaka}}, \bibinfo {author} {\bibfnamefont {K.}~\bibnamefont {Tsuda}}, \bibinfo {author} {\bibfnamefont {J.}~\bibnamefont {Shiomi}},\ and\ \bibinfo {author} {\bibfnamefont {R.}~\bibnamefont {Tamura}},\ }\bibfield  {title} {\bibinfo {title} {Designing metamaterials with quantum annealing and factorization machines},\ }\href@noop {} {\bibfield  {journal} {\bibinfo  {journal} {Physical Review Research}\ }\textbf {\bibinfo {volume} {2}},\ \bibinfo {pages} {013319} (\bibinfo {year} {2020})}\BibitemShut {NoStop}%
\bibitem [{\citenamefont {Tamura}\ \emph {et~al.}(2026)\citenamefont {Tamura}, \citenamefont {Seki}, \citenamefont {Minamoto}, \citenamefont {Kitai}, \citenamefont {Matsuda}, \citenamefont {Tanaka},\ and\ \citenamefont {Tsuda}}]{tamura2026black}%
  \BibitemOpen
  \bibfield  {author} {\bibinfo {author} {\bibfnamefont {R.}~\bibnamefont {Tamura}}, \bibinfo {author} {\bibfnamefont {Y.}~\bibnamefont {Seki}}, \bibinfo {author} {\bibfnamefont {Y.}~\bibnamefont {Minamoto}}, \bibinfo {author} {\bibfnamefont {K.}~\bibnamefont {Kitai}}, \bibinfo {author} {\bibfnamefont {Y.}~\bibnamefont {Matsuda}}, \bibinfo {author} {\bibfnamefont {S.}~\bibnamefont {Tanaka}},\ and\ \bibinfo {author} {\bibfnamefont {K.}~\bibnamefont {Tsuda}},\ }\bibfield  {title} {\bibinfo {title} {Black-box optimization using factorization and {Ising} machines},\ }\href@noop {} {\bibfield  {journal} {\bibinfo  {journal} {Applied Physics Reviews}\ }\textbf {\bibinfo {volume} {13}},\ \bibinfo {pages} {021307} (\bibinfo {year} {2026})}\BibitemShut {NoStop}%
\bibitem [{\citenamefont {Rendle}(2010)}]{rendle2010factorization}%
  \BibitemOpen
  \bibfield  {author} {\bibinfo {author} {\bibfnamefont {S.}~\bibnamefont {Rendle}},\ }\bibfield  {title} {\bibinfo {title} {Factorization machines},\ }in\ \href@noop {} {\emph {\bibinfo {booktitle} {2010 IEEE International conference on data mining}}}\ (\bibinfo {organization} {IEEE},\ \bibinfo {year} {2010})\ pp.\ \bibinfo {pages} {995--1000}\BibitemShut {NoStop}%
\bibitem [{\citenamefont {Kim}\ \emph {et~al.}(2022)\citenamefont {Kim}, \citenamefont {Shang}, \citenamefont {Moon}, \citenamefont {Pastega}, \citenamefont {Lee},\ and\ \citenamefont {Luo}}]{kim2022high}%
  \BibitemOpen
  \bibfield  {author} {\bibinfo {author} {\bibfnamefont {S.}~\bibnamefont {Kim}}, \bibinfo {author} {\bibfnamefont {W.}~\bibnamefont {Shang}}, \bibinfo {author} {\bibfnamefont {S.}~\bibnamefont {Moon}}, \bibinfo {author} {\bibfnamefont {T.}~\bibnamefont {Pastega}}, \bibinfo {author} {\bibfnamefont {E.}~\bibnamefont {Lee}},\ and\ \bibinfo {author} {\bibfnamefont {T.}~\bibnamefont {Luo}},\ }\bibfield  {title} {\bibinfo {title} {High-performance transparent radiative cooler designed by quantum computing},\ }\href@noop {} {\bibfield  {journal} {\bibinfo  {journal} {ACS Energy Letters}\ }\textbf {\bibinfo {volume} {7}},\ \bibinfo {pages} {4134} (\bibinfo {year} {2022})}\BibitemShut {NoStop}%
\bibitem [{\citenamefont {Kim}\ \emph {et~al.}(2024)\citenamefont {Kim}, \citenamefont {Jung}, \citenamefont {Bobbitt}, \citenamefont {Lee},\ and\ \citenamefont {Luo}}]{kim2024wide}%
  \BibitemOpen
  \bibfield  {author} {\bibinfo {author} {\bibfnamefont {S.}~\bibnamefont {Kim}}, \bibinfo {author} {\bibfnamefont {S.}~\bibnamefont {Jung}}, \bibinfo {author} {\bibfnamefont {A.}~\bibnamefont {Bobbitt}}, \bibinfo {author} {\bibfnamefont {E.}~\bibnamefont {Lee}},\ and\ \bibinfo {author} {\bibfnamefont {T.}~\bibnamefont {Luo}},\ }\bibfield  {title} {\bibinfo {title} {Wide-angle spectral filter for energy-saving windows designed by quantum annealing-enhanced active learning},\ }\href@noop {} {\bibfield  {journal} {\bibinfo  {journal} {Cell Reports Physical Science}\ }\textbf {\bibinfo {volume} {5}},\ \bibinfo {pages} {101891} (\bibinfo {year} {2024})}\BibitemShut {NoStop}%
\bibitem [{\citenamefont {Lin}\ \emph {et~al.}(2025)\citenamefont {Lin}, \citenamefont {Tada}, \citenamefont {Koizumi}, \citenamefont {Sumita}, \citenamefont {Tsuda},\ and\ \citenamefont {Tamura}}]{Lin2025ProtonIsingBBO}%
  \BibitemOpen
  \bibfield  {author} {\bibinfo {author} {\bibfnamefont {J.}~\bibnamefont {Lin}}, \bibinfo {author} {\bibfnamefont {T.}~\bibnamefont {Tada}}, \bibinfo {author} {\bibfnamefont {A.}~\bibnamefont {Koizumi}}, \bibinfo {author} {\bibfnamefont {M.}~\bibnamefont {Sumita}}, \bibinfo {author} {\bibfnamefont {K.}~\bibnamefont {Tsuda}},\ and\ \bibinfo {author} {\bibfnamefont {R.}~\bibnamefont {Tamura}},\ }\bibfield  {title} {\bibinfo {title} {Determination of stable proton configurations by black-box optimization using an {Ising} machine},\ }\href {https://doi.org/10.1021/acs.jpcc.4c07104} {\bibfield  {journal} {\bibinfo  {journal} {The Journal of Physical Chemistry C}\ }\textbf {\bibinfo {volume} {129}},\ \bibinfo {pages} {2332} (\bibinfo {year} {2025})}\BibitemShut {NoStop}%
\bibitem [{\citenamefont {Urushihara}\ \emph {et~al.}(2023)\citenamefont {Urushihara}, \citenamefont {Karube}, \citenamefont {Yamaguchi},\ and\ \citenamefont {Tamura}}]{urushihara2023optimization}%
  \BibitemOpen
  \bibfield  {author} {\bibinfo {author} {\bibfnamefont {M.}~\bibnamefont {Urushihara}}, \bibinfo {author} {\bibfnamefont {M.}~\bibnamefont {Karube}}, \bibinfo {author} {\bibfnamefont {K.}~\bibnamefont {Yamaguchi}},\ and\ \bibinfo {author} {\bibfnamefont {R.}~\bibnamefont {Tamura}},\ }\bibfield  {title} {\bibinfo {title} {Optimization of core--shell nanoparticles using a combination of machine learning and {Ising} machine},\ }\href@noop {} {\bibfield  {journal} {\bibinfo  {journal} {Advanced Photonics Research}\ }\textbf {\bibinfo {volume} {4}},\ \bibinfo {pages} {2300226} (\bibinfo {year} {2023})}\BibitemShut {NoStop}%
\bibitem [{\citenamefont {Nawa}\ \emph {et~al.}(2023)\citenamefont {Nawa}, \citenamefont {Suzuki}, \citenamefont {Masuda}, \citenamefont {Tanaka},\ and\ \citenamefont {Miura}}]{Nawa2023MTJQA}%
  \BibitemOpen
  \bibfield  {author} {\bibinfo {author} {\bibfnamefont {K.}~\bibnamefont {Nawa}}, \bibinfo {author} {\bibfnamefont {T.}~\bibnamefont {Suzuki}}, \bibinfo {author} {\bibfnamefont {K.}~\bibnamefont {Masuda}}, \bibinfo {author} {\bibfnamefont {S.}~\bibnamefont {Tanaka}},\ and\ \bibinfo {author} {\bibfnamefont {Y.}~\bibnamefont {Miura}},\ }\bibfield  {title} {\bibinfo {title} {Quantum annealing optimization method for the design of barrier materials in magnetic tunnel junctions},\ }\href {https://doi.org/10.1103/PhysRevApplied.20.024044} {\bibfield  {journal} {\bibinfo  {journal} {Physical Review Applied}\ }\textbf {\bibinfo {volume} {20}},\ \bibinfo {pages} {024044} (\bibinfo {year} {2023})}\BibitemShut {NoStop}%
\bibitem [{\citenamefont {Hida}\ \emph {et~al.}(2024)\citenamefont {Hida}, \citenamefont {Ikeda}, \citenamefont {Maruo}, \citenamefont {Sato},\ and\ \citenamefont {Yamazaki}}]{hida2024topology}%
  \BibitemOpen
  \bibfield  {author} {\bibinfo {author} {\bibfnamefont {M.}~\bibnamefont {Hida}}, \bibinfo {author} {\bibfnamefont {H.}~\bibnamefont {Ikeda}}, \bibinfo {author} {\bibfnamefont {A.}~\bibnamefont {Maruo}}, \bibinfo {author} {\bibfnamefont {M.}~\bibnamefont {Sato}},\ and\ \bibinfo {author} {\bibfnamefont {T.}~\bibnamefont {Yamazaki}},\ }\bibfield  {title} {\bibinfo {title} {Topology optimization of analog circuit design via global optimization using factorization machines with digital annealer},\ }\href@noop {} {\bibfield  {journal} {\bibinfo  {journal} {Journal of Advanced Mechanical Design, Systems, and Manufacturing}\ }\textbf {\bibinfo {volume} {18}},\ \bibinfo {pages} {JAMDSM0076} (\bibinfo {year} {2024})}\BibitemShut {NoStop}%
\bibitem [{\citenamefont {Tamura}\ \emph {et~al.}(2024)\citenamefont {Tamura}, \citenamefont {Nagata}, \citenamefont {Sodeyama}, \citenamefont {Nakamura}, \citenamefont {Tokuhira}, \citenamefont {Shibata}, \citenamefont {Hammura}, \citenamefont {Sugisawa}, \citenamefont {Kawamura}, \citenamefont {Tsurimoto} \emph {et~al.}}]{tamura2024machine}%
  \BibitemOpen
  \bibfield  {author} {\bibinfo {author} {\bibfnamefont {R.}~\bibnamefont {Tamura}}, \bibinfo {author} {\bibfnamefont {K.}~\bibnamefont {Nagata}}, \bibinfo {author} {\bibfnamefont {K.}~\bibnamefont {Sodeyama}}, \bibinfo {author} {\bibfnamefont {K.}~\bibnamefont {Nakamura}}, \bibinfo {author} {\bibfnamefont {T.}~\bibnamefont {Tokuhira}}, \bibinfo {author} {\bibfnamefont {S.}~\bibnamefont {Shibata}}, \bibinfo {author} {\bibfnamefont {K.}~\bibnamefont {Hammura}}, \bibinfo {author} {\bibfnamefont {H.}~\bibnamefont {Sugisawa}}, \bibinfo {author} {\bibfnamefont {M.}~\bibnamefont {Kawamura}}, \bibinfo {author} {\bibfnamefont {T.}~\bibnamefont {Tsurimoto}}, \emph {et~al.},\ }\bibfield  {title} {\bibinfo {title} {Machine learning prediction of the mechanical properties of injection-molded polypropylene through {X}-ray diffraction analysis},\ }\href@noop {} {\bibfield  {journal} {\bibinfo  {journal} {Science and Technology of Advanced Materials}\ }\textbf {\bibinfo {volume} {25}},\ \bibinfo {pages} {2388016} (\bibinfo
  {year} {2024})}\BibitemShut {NoStop}%
\bibitem [{\citenamefont {Matsumori}\ \emph {et~al.}(2022)\citenamefont {Matsumori}, \citenamefont {Taki},\ and\ \citenamefont {Kadowaki}}]{Matsumori2022QUBOStructural}%
  \BibitemOpen
  \bibfield  {author} {\bibinfo {author} {\bibfnamefont {T.}~\bibnamefont {Matsumori}}, \bibinfo {author} {\bibfnamefont {M.}~\bibnamefont {Taki}},\ and\ \bibinfo {author} {\bibfnamefont {T.}~\bibnamefont {Kadowaki}},\ }\bibfield  {title} {\bibinfo {title} {Application of {QUBO} solver using black-box optimization to structural design for resonance avoidance},\ }\href {https://doi.org/10.1038/s41598-022-16327-7} {\bibfield  {journal} {\bibinfo  {journal} {Scientific Reports}\ }\textbf {\bibinfo {volume} {12}},\ \bibinfo {pages} {12143} (\bibinfo {year} {2022})}\BibitemShut {NoStop}%
\bibitem [{\citenamefont {Takaki}\ \emph {et~al.}(2025)\citenamefont {Takaki}, \citenamefont {Asai}, \citenamefont {Katayama}, \citenamefont {Urano}, \citenamefont {Yonezawa},\ and\ \citenamefont {Kawaguchi}}]{takaki2025joint}%
  \BibitemOpen
  \bibfield  {author} {\bibinfo {author} {\bibfnamefont {K.}~\bibnamefont {Takaki}}, \bibinfo {author} {\bibfnamefont {Y.}~\bibnamefont {Asai}}, \bibinfo {author} {\bibfnamefont {S.}~\bibnamefont {Katayama}}, \bibinfo {author} {\bibfnamefont {K.}~\bibnamefont {Urano}}, \bibinfo {author} {\bibfnamefont {T.}~\bibnamefont {Yonezawa}},\ and\ \bibinfo {author} {\bibfnamefont {N.}~\bibnamefont {Kawaguchi}},\ }\bibfield  {title} {\bibinfo {title} {Joint black-box optimization of warehouse layout and worker assignment using quantum annealing and factorization machines},\ }in\ \href@noop {} {\emph {\bibinfo {booktitle} {2025 IEEE International Conference on Systems, Man, and Cybernetics (SMC)}}}\ (\bibinfo {organization} {IEEE},\ \bibinfo {year} {2025})\ pp.\ \bibinfo {pages} {4837--4844}\BibitemShut {NoStop}%
\bibitem [{\citenamefont {Kikuchi}\ and\ \citenamefont {Tanaka}(2026)}]{kikuchi2026factorization}%
  \BibitemOpen
  \bibfield  {author} {\bibinfo {author} {\bibfnamefont {S.}~\bibnamefont {Kikuchi}}\ and\ \bibinfo {author} {\bibfnamefont {S.}~\bibnamefont {Tanaka}},\ }\bibfield  {title} {\bibinfo {title} {Factorization machine with quadratic-optimization annealing for {RNA} inverse folding and evaluation of binary-integer encoding and nucleotide assignment},\ }\href@noop {} {\bibfield  {journal} {\bibinfo  {journal} {arXiv preprint arXiv:2602.16643}\ } (\bibinfo {year} {2026})}\BibitemShut {NoStop}%
\bibitem [{\citenamefont {Minamoto}\ and\ \citenamefont {Sakamoto}(2025)}]{Minamoto2025Black}%
  \BibitemOpen
  \bibfield  {author} {\bibinfo {author} {\bibfnamefont {Y.}~\bibnamefont {Minamoto}}\ and\ \bibinfo {author} {\bibfnamefont {Y.}~\bibnamefont {Sakamoto}},\ }\bibfield  {title} {\bibinfo {title} {A black-box optimization method with polynomial-based kernels and quadratic-optimization annealing},\ }\href@noop {} {\bibfield  {journal} {\bibinfo  {journal} {arXiv preprint arXiv:2501.04225}\ } (\bibinfo {year} {2025})}\BibitemShut {NoStop}%
\bibitem [{\citenamefont {Endo}\ and\ \citenamefont {Takahashi}(2025)}]{endo2025function}%
  \BibitemOpen
  \bibfield  {author} {\bibinfo {author} {\bibfnamefont {K.}~\bibnamefont {Endo}}\ and\ \bibinfo {author} {\bibfnamefont {K.~Z.}\ \bibnamefont {Takahashi}},\ }\bibfield  {title} {\bibinfo {title} {Function smoothing regularization for precision factorization machine annealing in continuous variable optimization problems},\ }\href@noop {} {\bibfield  {journal} {\bibinfo  {journal} {Physical Review Research}\ }\textbf {\bibinfo {volume} {7}},\ \bibinfo {pages} {013149} (\bibinfo {year} {2025})}\BibitemShut {NoStop}%
\bibitem [{\citenamefont {Nakano}\ \emph {et~al.}(2026)\citenamefont {Nakano}, \citenamefont {Seki}, \citenamefont {Kikuchi},\ and\ \citenamefont {Tanaka}}]{nakano2026swift}%
  \BibitemOpen
  \bibfield  {author} {\bibinfo {author} {\bibfnamefont {M.}~\bibnamefont {Nakano}}, \bibinfo {author} {\bibfnamefont {Y.}~\bibnamefont {Seki}}, \bibinfo {author} {\bibfnamefont {S.}~\bibnamefont {Kikuchi}},\ and\ \bibinfo {author} {\bibfnamefont {S.}~\bibnamefont {Tanaka}},\ }\bibfield  {title} {\bibinfo {title} {Swift-fmqa: Enhancing factorization machine with quadratic-optimization annealing via sliding window},\ }\href@noop {} {\bibfield  {journal} {\bibinfo  {journal} {IEEE Access}\ }\textbf {\bibinfo {volume} {14}},\ \bibinfo {pages} {10977} (\bibinfo {year} {2026})}\BibitemShut {NoStop}%
\bibitem [{\citenamefont {Abe}\ \emph {et~al.}(2026)\citenamefont {Abe}, \citenamefont {Yamashita},\ and\ \citenamefont {Tanaka}}]{abe2026effectiveness}%
  \BibitemOpen
  \bibfield  {author} {\bibinfo {author} {\bibfnamefont {T.}~\bibnamefont {Abe}}, \bibinfo {author} {\bibfnamefont {M.}~\bibnamefont {Yamashita}},\ and\ \bibinfo {author} {\bibfnamefont {S.}~\bibnamefont {Tanaka}},\ }\bibfield  {title} {\bibinfo {title} {Effectiveness of binary autoencoders for qubo-based optimization problems},\ }\href@noop {} {\bibfield  {journal} {\bibinfo  {journal} {arXiv preprint arXiv:2602.10037}\ } (\bibinfo {year} {2026})}\BibitemShut {NoStop}%
\bibitem [{\citenamefont {Hayashi}\ \emph {et~al.}(2026)\citenamefont {Hayashi}, \citenamefont {Seki}, \citenamefont {Terada}, \citenamefont {Mukasa}, \citenamefont {Kikuchi},\ and\ \citenamefont {Tanaka}}]{hayashi2026improving}%
  \BibitemOpen
  \bibfield  {author} {\bibinfo {author} {\bibfnamefont {T.}~\bibnamefont {Hayashi}}, \bibinfo {author} {\bibfnamefont {Y.}~\bibnamefont {Seki}}, \bibinfo {author} {\bibfnamefont {K.}~\bibnamefont {Terada}}, \bibinfo {author} {\bibfnamefont {Y.}~\bibnamefont {Mukasa}}, \bibinfo {author} {\bibfnamefont {S.}~\bibnamefont {Kikuchi}},\ and\ \bibinfo {author} {\bibfnamefont {S.}~\bibnamefont {Tanaka}},\ }\bibfield  {title} {\bibinfo {title} {Improving fmqa via initial training data design considering marginal bit coverage in one-hot encoding},\ }\href@noop {} {\bibfield  {journal} {\bibinfo  {journal} {arXiv preprint arXiv:2605.04825}\ } (\bibinfo {year} {2026})}\BibitemShut {NoStop}%
\bibitem [{\citenamefont {Tucs}\ \emph {et~al.}(2026)\citenamefont {Tucs}, \citenamefont {Tamura},\ and\ \citenamefont {Tsuda}}]{tucs2026factorization}%
  \BibitemOpen
  \bibfield  {author} {\bibinfo {author} {\bibfnamefont {A.}~\bibnamefont {Tucs}}, \bibinfo {author} {\bibfnamefont {R.}~\bibnamefont {Tamura}},\ and\ \bibinfo {author} {\bibfnamefont {K.}~\bibnamefont {Tsuda}},\ }\bibfield  {title} {\bibinfo {title} {Factorization machine with iterative quantum reverse annealing: A {Python} package for batch black-box optimization with reverse quantum annealing},\ }\href@noop {} {\bibfield  {journal} {\bibinfo  {journal} {Advanced Intelligent Discovery}\ ,\ \bibinfo {pages} {e202500231}} (\bibinfo {year} {2026})}\BibitemShut {NoStop}%
\bibitem [{\citenamefont {Hama}\ and\ \citenamefont {Kadowaki}(2026)}]{hama2026subsampling}%
  \BibitemOpen
  \bibfield  {author} {\bibinfo {author} {\bibfnamefont {Y.}~\bibnamefont {Hama}}\ and\ \bibinfo {author} {\bibfnamefont {T.}~\bibnamefont {Kadowaki}},\ }\bibfield  {title} {\bibinfo {title} {Subsampling factorization machine annealing},\ }\href@noop {} {\bibfield  {journal} {\bibinfo  {journal} {Physical Review Research}\ }\textbf {\bibinfo {volume} {8}},\ \bibinfo {pages} {013187} (\bibinfo {year} {2026})}\BibitemShut {NoStop}%
\bibitem [{\citenamefont {Inoue}\ \emph {et~al.}(2022)\citenamefont {Inoue}, \citenamefont {Seki}, \citenamefont {Tanaka}, \citenamefont {Togawa}, \citenamefont {Ishizaki},\ and\ \citenamefont {Noda}}]{Inoue2022PhotonicQA}%
  \BibitemOpen
  \bibfield  {author} {\bibinfo {author} {\bibfnamefont {T.}~\bibnamefont {Inoue}}, \bibinfo {author} {\bibfnamefont {Y.}~\bibnamefont {Seki}}, \bibinfo {author} {\bibfnamefont {S.}~\bibnamefont {Tanaka}}, \bibinfo {author} {\bibfnamefont {N.}~\bibnamefont {Togawa}}, \bibinfo {author} {\bibfnamefont {K.}~\bibnamefont {Ishizaki}},\ and\ \bibinfo {author} {\bibfnamefont {S.}~\bibnamefont {Noda}},\ }\bibfield  {title} {\bibinfo {title} {Towards optimization of photonic-crystal surface-emitting lasers via quantum annealing},\ }\href {https://doi.org/10.1364/OE.471585} {\bibfield  {journal} {\bibinfo  {journal} {Optics Express}\ }\textbf {\bibinfo {volume} {30}},\ \bibinfo {pages} {43503} (\bibinfo {year} {2022})}\BibitemShut {NoStop}%
\bibitem [{\citenamefont {Kondo}\ \emph {et~al.}(2025)\citenamefont {Kondo}, \citenamefont {Kohira},\ and\ \citenamefont {Minamoto}}]{Kondo2025CarBodyFMQA}%
  \BibitemOpen
  \bibfield  {author} {\bibinfo {author} {\bibfnamefont {T.}~\bibnamefont {Kondo}}, \bibinfo {author} {\bibfnamefont {T.}~\bibnamefont {Kohira}},\ and\ \bibinfo {author} {\bibfnamefont {Y.}~\bibnamefont {Minamoto}},\ }\bibfield  {title} {\bibinfo {title} {Simultaneous structure design optimization of multiple car models using {FMQA}},\ }\href@noop {} {\bibfield  {journal} {\bibinfo  {journal} {Transactions of the Society of Automotive Engineers of Japan}\ }\textbf {\bibinfo {volume} {56}},\ \bibinfo {pages} {229} (\bibinfo {year} {2025})}\BibitemShut {NoStop}%
\bibitem [{\citenamefont {Kikuchi}\ \emph {et~al.}(2024)\citenamefont {Kikuchi}, \citenamefont {Takahashi},\ and\ \citenamefont {Tanaka}}]{kikuchi2024domainwall}%
  \BibitemOpen
  \bibfield  {author} {\bibinfo {author} {\bibfnamefont {S.}~\bibnamefont {Kikuchi}}, \bibinfo {author} {\bibfnamefont {K.}~\bibnamefont {Takahashi}},\ and\ \bibinfo {author} {\bibfnamefont {S.}~\bibnamefont {Tanaka}},\ }\bibfield  {title} {\bibinfo {title} {Performance of {Domain-Wall} encoding in a digital {Ising} machine},\ }\href@noop {} {\bibfield  {journal} {\bibinfo  {journal} {arXiv preprint arXiv:2410.11198}\ } (\bibinfo {year} {2024})}\BibitemShut {NoStop}%
\bibitem [{\citenamefont {Chen}\ \emph {et~al.}(2021)\citenamefont {Chen}, \citenamefont {Stollenwerk},\ and\ \citenamefont {Chancellor}}]{chen2021performance}%
  \BibitemOpen
  \bibfield  {author} {\bibinfo {author} {\bibfnamefont {J.}~\bibnamefont {Chen}}, \bibinfo {author} {\bibfnamefont {T.}~\bibnamefont {Stollenwerk}},\ and\ \bibinfo {author} {\bibfnamefont {N.}~\bibnamefont {Chancellor}},\ }\bibfield  {title} {\bibinfo {title} {Performance of {Domain-Wall} encoding for quantum annealing},\ }\href@noop {} {\bibfield  {journal} {\bibinfo  {journal} {IEEE Transactions on Quantum Engineering}\ }\textbf {\bibinfo {volume} {2}},\ \bibinfo {pages} {1} (\bibinfo {year} {2021})}\BibitemShut {NoStop}%
\bibitem [{\citenamefont {Tamura}\ \emph {et~al.}(2021)\citenamefont {Tamura}, \citenamefont {Shirai}, \citenamefont {Katsura}, \citenamefont {Tanaka},\ and\ \citenamefont {Togawa}}]{tamura2021performance}%
  \BibitemOpen
  \bibfield  {author} {\bibinfo {author} {\bibfnamefont {K.}~\bibnamefont {Tamura}}, \bibinfo {author} {\bibfnamefont {T.}~\bibnamefont {Shirai}}, \bibinfo {author} {\bibfnamefont {H.}~\bibnamefont {Katsura}}, \bibinfo {author} {\bibfnamefont {S.}~\bibnamefont {Tanaka}},\ and\ \bibinfo {author} {\bibfnamefont {N.}~\bibnamefont {Togawa}},\ }\bibfield  {title} {\bibinfo {title} {Performance comparison of typical binary-integer encodings in an {I}sing machine},\ }\href@noop {} {\bibfield  {journal} {\bibinfo  {journal} {IEEE Access}\ }\textbf {\bibinfo {volume} {9}},\ \bibinfo {pages} {81032} (\bibinfo {year} {2021})}\BibitemShut {NoStop}%
\bibitem [{\citenamefont {Seki}\ \emph {et~al.}(2022)\citenamefont {Seki}, \citenamefont {Tamura},\ and\ \citenamefont {Tanaka}}]{seki2022black}%
  \BibitemOpen
  \bibfield  {author} {\bibinfo {author} {\bibfnamefont {Y.}~\bibnamefont {Seki}}, \bibinfo {author} {\bibfnamefont {R.}~\bibnamefont {Tamura}},\ and\ \bibinfo {author} {\bibfnamefont {S.}~\bibnamefont {Tanaka}},\ }\bibfield  {title} {\bibinfo {title} {{Black-box optimization for integer-variable problems using {Ising} machines and factorization machines}},\ }\href@noop {} {\bibfield  {journal} {\bibinfo  {journal} {arXiv preprint arXiv:2209.01016}\ } (\bibinfo {year} {2022})}\BibitemShut {NoStop}%
\bibitem [{\citenamefont {Rastrigin}(1974)}]{rastrigin1974systems}%
  \BibitemOpen
  \bibfield  {author} {\bibinfo {author} {\bibfnamefont {L.~A.}\ \bibnamefont {Rastrigin}},\ }\href@noop {} {\emph {\bibinfo {title} {Systems of Extremal Control}}}\ (\bibinfo  {publisher} {Mir},\ \bibinfo {address} {Moscow},\ \bibinfo {year} {1974})\BibitemShut {NoStop}%
\bibitem [{\citenamefont {Rudolph}(1990)}]{rudolph1990evolution}%
  \BibitemOpen
  \bibfield  {author} {\bibinfo {author} {\bibfnamefont {G.}~\bibnamefont {Rudolph}},\ }\emph {\bibinfo {title} {Globale Optimierung mit parallelen Evolutionsstrategien}},\ \href@noop {} {\bibinfo {type} {Diplomarbeit}},\ \bibinfo  {school} {Department of Computer Science, University of Dortmund} (\bibinfo {year} {1990})\BibitemShut {NoStop}%
\bibitem [{\citenamefont {Loshchilov}\ and\ \citenamefont {Hutter}(2017)}]{loshchilov2017decoupled}%
  \BibitemOpen
  \bibfield  {author} {\bibinfo {author} {\bibfnamefont {I.}~\bibnamefont {Loshchilov}}\ and\ \bibinfo {author} {\bibfnamefont {F.}~\bibnamefont {Hutter}},\ }\bibfield  {title} {\bibinfo {title} {Decoupled weight decay regularization},\ }\href@noop {} {\bibfield  {journal} {\bibinfo  {journal} {arXiv preprint arXiv:1711.05101}\ } (\bibinfo {year} {2017})}\BibitemShut {NoStop}%
\bibitem [{\citenamefont {Demo}\ \emph {et~al.}(2021)\citenamefont {Demo}, \citenamefont {Tezzele},\ and\ \citenamefont {Rozza}}]{demo2021supervised}%
  \BibitemOpen
  \bibfield  {author} {\bibinfo {author} {\bibfnamefont {N.}~\bibnamefont {Demo}}, \bibinfo {author} {\bibfnamefont {M.}~\bibnamefont {Tezzele}},\ and\ \bibinfo {author} {\bibfnamefont {G.}~\bibnamefont {Rozza}},\ }\bibfield  {title} {\bibinfo {title} {A supervised learning approach involving active subspaces for an efficient genetic algorithm in high-dimensional optimization problems},\ }\href@noop {} {\bibfield  {journal} {\bibinfo  {journal} {SIAM Journal on Scientific Computing}\ }\textbf {\bibinfo {volume} {43}},\ \bibinfo {pages} {B831} (\bibinfo {year} {2021})}\BibitemShut {NoStop}%
\bibitem [{\citenamefont {{Fixstars Amplify}}(2021)}]{FixstarsAmplify}%
  \BibitemOpen
  \bibfield  {author} {\bibinfo {author} {\bibnamefont {{Fixstars Amplify}}},\ }\href {https://amplify.fixstars.com/en/} {\bibinfo {title} {{Fixstars Amplify Annealing Engine}}} (\bibinfo {year} {2021}),\ \bibinfo {note} {https://amplify.fixstars.com/en/}\BibitemShut {NoStop}%
\bibitem [{\citenamefont {Akiba}\ \emph {et~al.}(2019)\citenamefont {Akiba}, \citenamefont {Sano}, \citenamefont {Yanase}, \citenamefont {Ohta},\ and\ \citenamefont {Koyama}}]{akiba2019optuna}%
  \BibitemOpen
  \bibfield  {author} {\bibinfo {author} {\bibfnamefont {T.}~\bibnamefont {Akiba}}, \bibinfo {author} {\bibfnamefont {S.}~\bibnamefont {Sano}}, \bibinfo {author} {\bibfnamefont {T.}~\bibnamefont {Yanase}}, \bibinfo {author} {\bibfnamefont {T.}~\bibnamefont {Ohta}},\ and\ \bibinfo {author} {\bibfnamefont {M.}~\bibnamefont {Koyama}},\ }\bibfield  {title} {\bibinfo {title} {Optuna: A next-generation hyperparameter optimization framework},\ }in\ \href@noop {} {\emph {\bibinfo {booktitle} {Proceedings of the 25th ACM SIGKDD international conference on knowledge discovery \& data mining}}}\ (\bibinfo {year} {2019})\ pp.\ \bibinfo {pages} {2623--2631}\BibitemShut {NoStop}%
\bibitem [{\citenamefont {Rosenbrock}(1960)}]{rosenbrock1960automatic}%
  \BibitemOpen
  \bibfield  {author} {\bibinfo {author} {\bibfnamefont {H.~H.}\ \bibnamefont {Rosenbrock}},\ }\bibfield  {title} {\bibinfo {title} {An automatic method for finding the greatest or least value of a function},\ }\href@noop {} {\bibfield  {journal} {\bibinfo  {journal} {The Computer Journal}\ }\textbf {\bibinfo {volume} {3}},\ \bibinfo {pages} {175} (\bibinfo {year} {1960})}\BibitemShut {NoStop}%
\end{thebibliography}%
\end{document}